\documentclass{article}

\usepackage[preprint,nonatbib]{neurips_2025}

\usepackage[utf8]{inputenc} 
\usepackage[T1]{fontenc}    
\usepackage{hyperref}       
\usepackage{url}            
\usepackage{booktabs}       
\usepackage{amsfonts}       
\usepackage{nicefrac}       
\usepackage{microtype}      
\usepackage{xcolor}         
\usepackage{multirow}
\usepackage{wrapfig}
\usepackage{graphicx}

\usepackage[most]{tcolorbox}

\definecolor{titlebg}{RGB}{85,85,85}      
\definecolor{bodybg}{RGB}{240,240,240}    
\usepackage{color}
\usepackage{amsmath}
\usepackage{amssymb}
\usepackage{amsthm}
\usepackage{caption}
\usepackage{subcaption}
\usepackage{mathtools}
\usepackage{appendix}
\usepackage{tabularx,colortbl}

\usepackage{abstract}

\theoremstyle{definition}

\definecolor{lightGreen}{RGB}{150,255,150}

\title{Patho-R1: A Multimodal Reinforcement Learning-Based Pathology Expert Reasoner}

\author{
\textbf{Wenchuan Zhang}$^{1,2}$\thanks{Equal contribution.} \quad
\textbf{Penghao Zhang}$^{3}$\footnotemark[1] \quad
\textbf{Jingru Guo}$^{4}$\footnotemark[1] \quad
\textbf{Tao Cheng}$^{5}$ \quad
\textbf{Jie Chen}$^{2}$ \quad \\
\textbf{Shuwan Zhang}$^{6}$ \quad 
\textbf{Zhang Zhang}$^{1}$ \quad
\textbf{Yuhao Yi}$^{1,2}$\thanks{Corresponding author.} \quad
\textbf{Hong Bu}$^{1,2}$ \\
\\
$^1$Department of Pathology, West China Hospital, Sichuan University\\
$^2$Institute of Clinical Pathology, West China Hospital, Sichuan University\\
$^3$Independent Researcher \quad $^4$University of Toronto \quad $^5$Business School, Sichuan University\\
$^6$Department of Pathology, Shengjing Hospital of China Medical University\\
\texttt{zhangwenchuan@stu.scu.edu.cn, yuhaoyi@scu.edu.cn}
}

\begin{document}

\maketitle

\begin{abstract}
  Recent advances in vision-language models (VLMs) have enabled broad progress in the general medical field. However, pathology still remains a more challenging sub-domain, with current pathology-specific VLMs exhibiting limitations in both diagnostic accuracy and reasoning plausibility. Such shortcomings are largely attributable to the nature of current pathology datasets, which are primarily composed of image–description pairs that lack the depth and structured diagnostic paradigms employed by real-world pathologists. In this study, we leverage pathology textbooks and real-world pathology experts to construct high-quality, reasoning-oriented datasets. Building on this, we introduce Patho-R1, a multimodal RL-based pathology Reasoner, trained through a three-stage pipeline: (1) continued pretraining on 3.5 million image-text pairs for knowledge infusion; (2) supervised fine-tuning on 500k high-quality Chain-of-Thought samples for reasoning incentivizing; (3) reinforcement learning using Group Relative Policy Optimization and Decoupled Clip and Dynamic sAmpling Policy Optimization strategies for multimodal reasoning quality refinement. To further assess the alignment quality of our dataset, we propose Patho-CLIP, trained on the same figure-caption corpus used for continued pretraining. Comprehensive experimental results demonstrate that both Patho-CLIP and Patho-R1 achieve robust performance across a wide range of pathology-related tasks, including zero-shot classification, cross-modal retrieval, Visual Question Answering, and Multiple Choice Question. Our project is available at the Patho-R1 repository: 
  \url{https://github.com/Wenchuan-Zhang/Patho-R1}.

\end{abstract}

\begin{figure}[h]
    \centering
    \hspace*{1.1em}
    \includegraphics[width=0.96\linewidth]{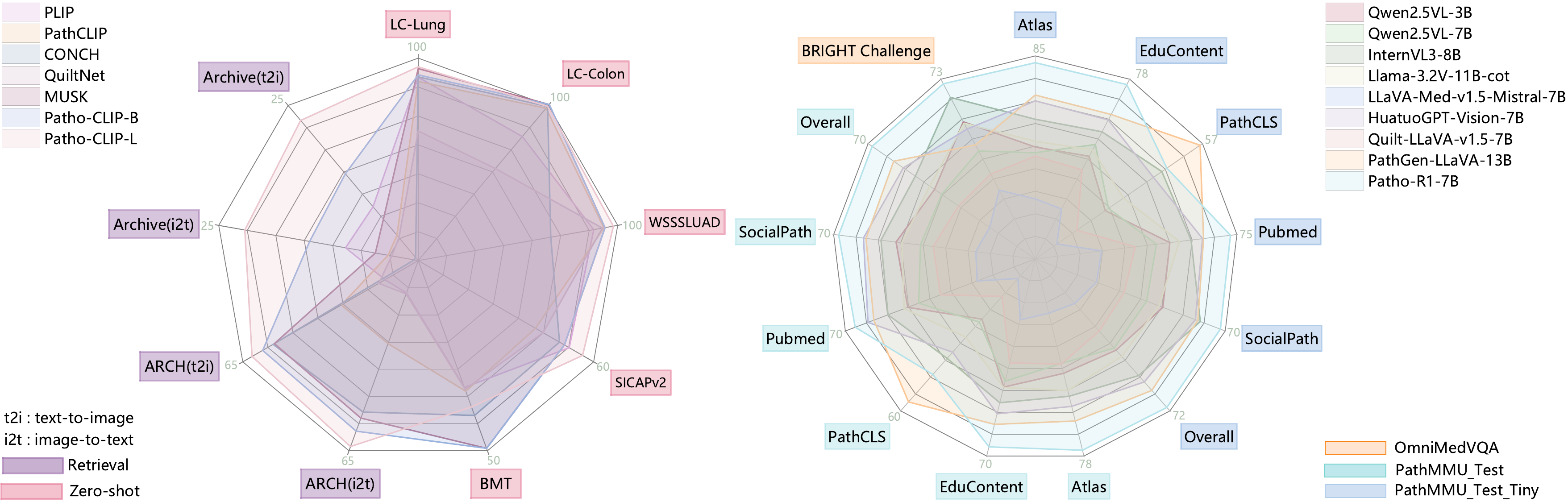}
    \caption{The performance of the proposed Patho-CLIP (left), and the Patho-R1 (right).}
    \label{fig:1}
\end{figure}

\section{Introduction}

In the medical domain, pathology, as the gold standard of modern clinical diagnosis \cite{tolkach2023artificial}, plays a pivotal role in guiding therapeutic decisions \cite{foersch2023multistain}. However, building robust AI systems for pathology presents significantly greater challenges compared to other medical imaging tasks like MRI \cite{wang2025triad} or CT \cite{wu2024voco}. This stems from the inherent complexity and fine-grained nature of pathology data, coupled with the scarcity of high-quality, expert-annotated multimodal datasets \cite{xie2024medtrinity}. While general vision-language models (VLMs), such as CLIP \cite{CLIP} and LLaVA \cite{LLAVA}, have shown promising results across various medical imaging tasks \cite{medclip,medflamingo,radfm}, their performance in pathology remains limited---especially in terms of cross-disease generalization and the comprehension of complex diagnostic tasks---highlighting the urgent need for pathology-specific solutions \cite{pmcclip,biomedgpt,llavamed}.

To address this gap, recent efforts \cite{quiltllava, pathgen16m} introduced pathology-adapted multimodal datasets \cite{quilt1m, plip, pathgen16m}, demonstrating impressive performance on various benchmark tasks, facilitating the initial transformation of generic multimodal architectures to the pathology domain. However, despite their competitive results on constrained tasks such as Multiple Choice Question (MCQ) and Visual Question Answering (VQA), the underlying decision-making principles of the models remain opaque, limiting their applicability in real-world clinical settings, where interpretability and trustworthiness are critical for deployment. Existing datasets, constructed from educational videos, social media posts, and institutional archives share a common characteristic: the pairing between visual and textual modalities tends to privilege surface-level descriptions over semantically rich, diagnostically grounded interpretations. ``Textbook-quality'' data \cite{gunasekar2023textbooksneed}, encompassing detailed disease mechanisms and organized domain knowledge, may predictably yield a model's capacity for clinically grounded reasoning.

Notably, recent success in reinforcement learning, exemplified by DeepSeek-R1 \cite{deepseekr1}, QwQ-32B \cite{qwq32b}, Kimi k1.5 \cite{kimik15}, have revealed the strong potential of reinforcement learning in fostering reasoning abilities within language models. Among existing methods, the Group Relative Policy Optimization (GRPO) stands out for its simple but effective ground truth-driven reward design. Numerous subsequent works continuously validate its capability, not only in logic-intensive tasks like math and coding, but in domains that, while less dependent on symbolic deduction, require systematic thinking and structured judgement instead \cite{visionr1incentivizing, medr1, medvlmr1, finr1}. More intriguingly, recent works have pushed GRPO forward: the Decoupled Clip and Dynamic sAmpling Policy Optimization (DAPO) \cite{Dapo} leverages higher clip ratio and dynamically sampled training batches to yield promising performance with fewer training steps, offering an inspiring alternative to domain-aligned reward mechanism design.

Specifically, our key contributions include: 
\begin{itemize}
    \item We propose a comprehensive \textbf{data curation pipeline} that requires \textbf{minimal human effort} while ensuring scalable generation of \textbf{high-quality SFT data with reasoning}. 
    \item We present \textbf{\textit{Patho-CLIP}}, an open-source pathology adapted CLIP model, which outperforms state-of-the-art models in classification and retrieval tasks. 
    \item We explore the end-to-end training process of domain adaptation of pretrained vision-language models, especially the latest reinforcement learning methods: \textbf{GRPO} and \textbf{DAPO}. We release the model weights of our vision-language pathology reasoning model: \textbf{\textit{Patho-R1}}, which demonstrates superior performance on various benchmarks as shown in Figure~\ref{fig:1}.
\end{itemize}

\section{Related Work}

\paragraph{Large Vision Language Models in Medical Fields.} The emergence of general-purpose VLMs such as CLIP \cite{CLIP} has significantly advanced the development of multimodal models in the medical domain. To better capture domain-specific semantics, various CLIP-based adaptations have been proposed, including PubMedCLIP \cite{pubmedclip}, MedCLIP \cite{medclip}, BiomedCLIP \cite{BiomedCLIP}, PMC-CLIP \cite{pmcclip}, UniMed-CLIP \cite{unimedclip}, and Med-PaLM \cite{medpalm}. These models leverage large-scale medical image-text pairs for contrastive pretraining, leading to improved performance on tasks such as medical image classification and retrieval. To further enhance the utility of medical VLMs, a new line of multimodal medical assistants has emerged, including BiomedGPT \cite{biomedgpt}, Med-Flamingo \cite{medflamingo}, LLaVA-Med \cite{llavamed}, RadFM \cite{radfm}, and HuatuoGPT-Vision \cite{huatuogptvision}. These models integrate large language models to enable image-conditioned dialogue and question answering, aiming to support more interactive and context-aware medical understanding. In the field of pathology, where challenges such as extremely high-resolution images and subtle morphological distinctions prevail, several domain-specific CLIP-structured models have been developed, including PLIP \cite{plip}, CONCH \cite{conch}, MUSK \cite{musk}, TITAN \cite{titan}, and QuiltNet \cite{quilt1m}. Alongside these, a growing number of pathology-focused multimodal models have been introduced, such as PathAsst \cite{pathasst}, Quilt-LLaVA \cite{quiltllava}, PathChat \cite{pathchat}, CPath-Omni\cite{cpathomni}, PathGen-LLaVA \cite{pathgen16m}, and PA-LLaVA \cite{pallava}.
These models attempt to adapt general-purpose architectures to the pathology domain and incorporate capabilities such as visual question answering and multi-turn dialogue to facilitate diagnostic support.

\paragraph{Vision-Language Reinforcement Learning for Post-Training.} Recently, the focus of large language model (LLM) research has been shifting from traditional supervised learning to reinforcement learning (RL) as a means to enhance reasoning capabilities \cite{deepseekr1, openaio1}. 
One line of work centers around Chain-of-Thought (CoT) prompting: it involves constructing structured multimodal reasoning datasets and performing instruction tuning to improve the model’s reasoning traceability, as seen in LLaVA-CoT \cite{llavacot} and LlamaV-o1 \cite{llamavo1}. In contrast, another class of approaches, such as Visual-RFT \cite{visualrft}, VLM-R1 \cite{vlmr1} , and Perception-R1 \cite{perceptionr1}, bypasses CoT-style instruction tuning and instead focuses on leveraging RL to directly enhance visual perception. These methods typically rely on carefully designed reward functions to guide models toward learning more robust visual representations under unsupervised or weakly supervised conditions. A further evolution of this line of work is exemplified by R1-OneVision \cite{r1onevision}, R1-V \cite{r1v}, and Vision-R1 \cite{visionr1incentivizing}, which propose a novel cross-modal reasoning pipeline. These methods first convert images into formalized language representations that encapsulate visual semantics, then construct language-centric visual reasoning datasets for instruction tuning, followed by an RL phase to incentivize deeper multimodal reasoning. Although these methods have made significant progress in general fields, their exploration in the field of medical imaging is still in its early stages. For example, MedVLM-R1 \cite{medvlmr1} and Med-R1 \cite{medr1} have only made preliminary adaptations to medical scenarios and lack system modeling and expert knowledge guidance for specific tasks. 


\section{Method}
\label{sec:method}
\begin{figure}[t]
  \centering
  \includegraphics[width=1.0\textwidth]{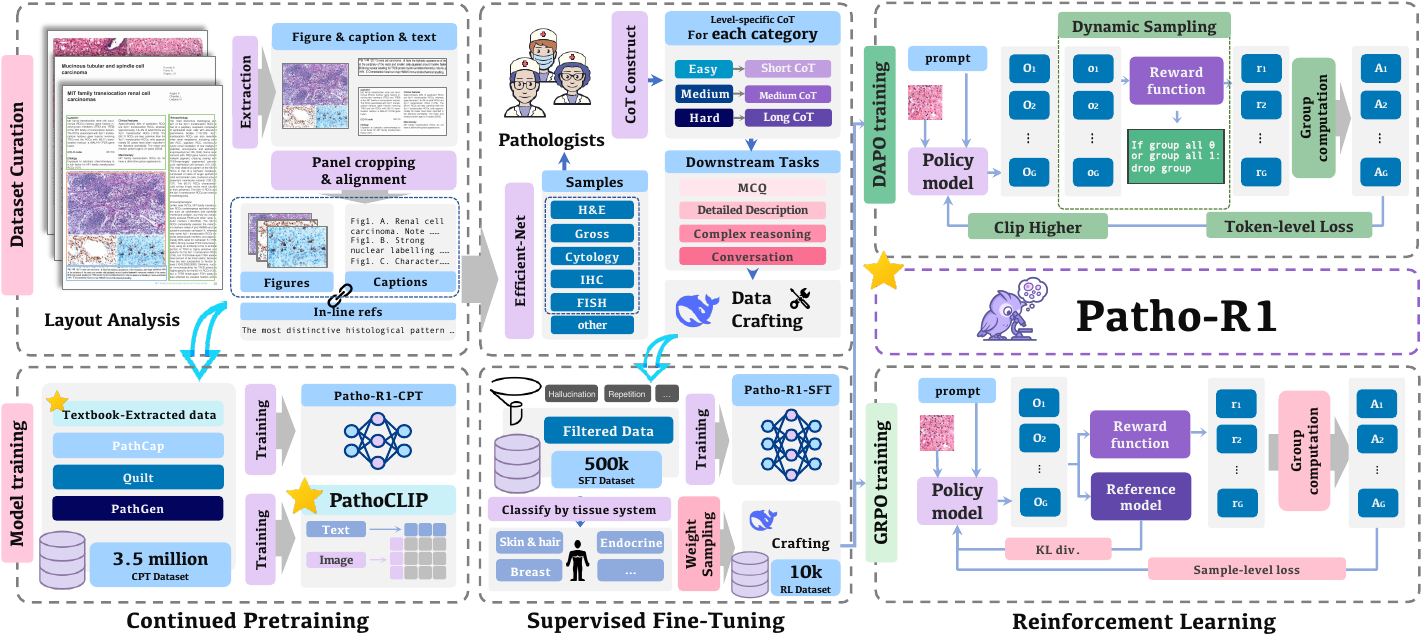}
  \caption{The sequential pipeline for developing Patho-R1. We start by extracting and aligning figures, captions, and in-line references. The image caption pairs, combined with public datasets, will be used for Patho-CLIP training and CPT for Patho-R1 training. Pairs with in-line references will be categorized into pathological sub-fields, clustered by difficulty, applied into tasks with level-specific CoTs to construct SFT data. Data with diagnosis information will be sampled for the final RL.}
  \label{fig:pipeline}
\end{figure}

\subsection{Dataset Overview}

As the potential of VLMs for domain-specific applications continues to be explored, constructing high-quality datasets is increasingly crucial for developing clinically meaningful models. However, existing datasets have two main drawbacks: (1) \textit{Existing datasets most focus on general medical domains, leaving a gap in the collection of pathology-specific data;} (2) \textit{They often lack detailed, authoritative, and diagnostically accurate records of the diagnostic process, focusing primarily on simple descriptions and conclusions.} To address these issues, we curated a large-scale, pathology-specific multimodal corpus covering 3 publicly available datasets, 
660 authoritative pathology textbooks and education notes. The overview of our data curation and model training pipeline is shown in Figure~\ref{fig:pipeline}. Datasets used in each training phase are described below. For more details of the training data, see Appendix~\ref{appendix:Datasets construction}.

\textbf{Continued Pretraining (CPT):} Our CPT dataset comprises a total of \textbf{3.5 million image-text pairs}, of which 2.8 million are from PubMed \cite{pathasst}, Quilt \cite{quilt1m}, PathGen \cite{pathgen16m}, and 0.7 million from pathology textbooks and notes. This dataset also serves as the training data for our PathoCLIP.

\textbf{Supervised Fine-tuning (SFT):} Our SFT dataset comprises \textbf{500k samples} drawn from 5 pathological subfields: histopathology, gross examination, immunohistochemistry (IHC), cytology, fluorescence in situ hybridization (FISH); with 3 distinct levels of Chain-of-Thoughts per subfield; spanning 4 downstream task types: descriptive analysis, complex reasoning, multi-turn conversations, multiple-choice questions. This results in a total of 60 data combination types.

\textbf{Reinforcement Learning:} Our RL dataset comprises 10k diagnostic-oriented MCQs, aligned with the same five pathological subfields defined in the SFT stage.

\subsection{Continued Pretraining and PathoCLIP Training}

\textbf{Data Preparation.} We first scanned pathology textbooks into high-resolution images and used DocLayoutYolo \cite{zhao2024doclayoutyoloenhancingdocumentlayout} to segment each page into body text, figure panels, and captions. Figure-caption pairs were matched based on spatial layout, and captions were extracted using OCR. For multi-panel figures, we applied edge detection and label recognition to split composite images and align sub-figures with corresponding caption segments. To leverage the rich contextual information in textbook narratives, we also extracted body text via OCR and used the Qwen-max model to automatically identify in-line references between figures and surrounding text, reducing the need for manual annotation. Further details are provided in Appendix~\ref{appendix:image-text pair sources}.

\textbf{PathoCLIP Training.} We employ OpenAI-CLIP-B and OpenAI-CLIP-L as backbone architectures of our model. To construct a cross-modal model endowed with both fine-grained morphological representation and clinical pathology comprehension capabilities, we implement a two-stage progressive training paradigm inspired by \cite{pathgen16m}: in stage I, we initialize the model via contrastive learning on the PathGen-1.6M dataset, which emphasizes tissue-cell morphology and spatial organization to instill high-resolution morphological priors; in stage II, we preserve these morphological representations while integrating PathGen-1.6M with Quilt-1M, PathCap, and our textbook-extracted dataset, yielding a composite corpus of 3.5 million image–text pairs. Empirical evaluations demonstrate that this progressive strategy not only significantly enhances the model’s ability to discriminate tissue heterogeneity but achieves state-of-the-art semantic understanding and image–text retrieval performance on a range of pathological diagnostic benchmarks. For more details on PathoCLIP training, see Appendix~\ref{appendix:clip training}.

\textbf{Patho-R1 Continued Pretraining.} Continued Pretraining has been empirically validated to be an effective approach to inject domain-specific knowledge into pretrained language models \cite{parmar2024reuse}, as well as achieve competitive or even superior performance compared to larger-scale models \cite{huatuogptvision}. To ensure a wide coverage of pathological knowledge, we incorporated all publicly available pathology datasets with our retrieved context-rich image-caption pairs. Given that the total dataset contains 3.5 million samples, we selected Qwen2.5VL-3B and Qwen2.5VL-7B as our base models to balance model capacity and data volume.

\subsection{Invoking Reasoning Ability via SFT}

Continued pretraining on domain-specific corpora can endow the model with domain knowledge, but hurt its instruction following ability \cite{cheng2024adaptinglargelanguagemodels}. During this phase, our intentions are: (1) \textit{recovering the model's instruction-following ability;} (2) \textit{invoking reasoning behavior tailored to pathological diagnostics.} Unlike general-domain data, pathological diagnosis is highly structured---real-world pathologists typically follow a systematic process when interpreting histological images. Hence, a cost-effective data generation strategy that balances data granularity with diagnostic specificity is crucial for this training phase.

We started by categorizing textbook-extracted image-caption pairs into pathology subfields. A small subset of data was manually labeled to serve as supervision for training an EfficientNet-based \cite{efficientnet} classifier. Batched inference results on the unlabeled data were manually inspected and corrected before being incorporated into the training set to iteratively boost classifier training. Among the 10 initially identified subfields, we chose 5 with the largest data volume and highest pathological relevance as candidate subfields for SFT: histopathology, gross examination, IHC, cytology and FISH. However, five subfields were still far from meeting the requirement of data diversity. To address this, inspired by \cite{autocluster}, we performed a three-way K-means clustering within each subfield based on the joint embeddings of images and captions. Interestingly, we observed consistent difficulty-level distinctions among the resulting clusters, based on which we designed 3 levels of CoT reasoning: easy, medium and hard. Finally, we combined the 15 distinct CoTs with 4 types of downstream tasks: MCQ, detailed description, complex reasoning, and multi-turn conversation, yielding a total of 60 prompts for SFT data generation. For more details on SFT data generation, see Appendix~\ref{appendix:cot-sft data generation scheme}.

We selected DeepSeek-R1 as our SFT data generation model, primarily for the following two reasons. First, as existing VLMs have not been adapted to the pathology domain, their performance in pathological imagery understanding is not satisfactory. In our experiments, providing VLMs with image-caption-CoT prompt triplets did not yield better results compared to using only captions with a strong text-only model. Second, DeepSeek-R1, as a state-of-the-art reasoning model, shows strong capabilities in multi-step thinking and long-context understanding---both of which are crucial for handling our lengthy and complex CoT-style prompts. However, this choice inevitably suffered from certain issues due to DeepSeek's inherent weaknesses, such as language mixing in pathological terminologies and endless repetition. To mitigate generation noise, we implemented a quality control stage involving rule-based filtering and selective manual verification. The final cleaned dataset comprises 500k samples. For implementation details, see Appendix~\ref{appendix:lvlm training}.

\subsection{Reinforcement Learning}

For reinforcement learning data generation, we constructed a diagnosis-oriented MCQ dataset by selecting a non-overlapping subset of samples from the SFT data. To ensure diversity, we performed proportionate stratified sampling based on subfield distribution. For the two largest subfields that can be further categorized by tissue systems, namely histopathology and gross examination, we divided the data accordingly and performed proportional sampling within each subfield. These samples were paired with diagnosis information to generate diagnosis-oriented MCQs. For more details on data generation, see Appendix~\ref{appendix:rl dataset construction}.

For each training iteration, GRPO and DAPO sample $G$ candidate output $\{o_i\}_{i=1}^G$ from the old policy $\pi_{\mathrm{old}}$, then maximize the following objectives respectively and optimize the model $\pi_{\mathrm{\theta}}$: 

\begin{align}
\mathcal{J}_{\mathrm{GRPO}}(\theta) &= \mathbb{E}_{\mathbf{v}\sim P(\mathbf{V}),\,\{o_{\mathbf{i}}\}_{\mathbf{i}=1}^{G}\sim\pi_{\theta_{\mathbf{old}}}(O|\mathbf{v})} \nonumber \\
&\quad \left[\frac{1}{G}\sum_{i=1}^{G}\!\frac{1}{|o_i|}\!\sum_{t=1}^{|o_i|}\!
\left(
\min\left(r_{i,t}A_{i,t},\,\mathrm{clip}\left(r_{i,t},1\!-\!\epsilon,1\!+\!\epsilon\right)A_{i,t}\right)\!
-\! \beta D_{\mathrm{KL}}(\pi_{\theta}\parallel\pi_{\mathrm{ref}})
\right)\right] \\
\mathcal{J}_{\mathrm{DAPO}}(\theta) &= \mathbb{E}_{\mathbf{v}\sim P(\mathbf{V}),\,\{o_{\mathbf{i}}\}_{\mathbf{i}=1}^{G}\sim\pi_{\theta_{\mathbf{old}}}(O|\mathbf{v})} \nonumber \\
&\quad \left[\frac{1}{\sum_{i=1}^{G}|o_i|}\sum_{i=1}^{G}\sum_{t=1}^{|o_i|}
\min\left(r_{i,t}A_{i,t},\,\mathrm{clip}\left(r_{i,t},1-\epsilon_{\mathrm{low}},1+\epsilon_{\mathrm{high}}\right)A_{i,t}\right)\right] \nonumber \\
&\quad \text{s.t. } 0< \Big|\{o_i\mid\texttt{is\_equivalent}(a,o_i)\}\Big|< G.
\end{align}

where $A_{i}=\nicefrac{(r_{i}-\mu_{\{r\}})}{\sigma_{\{r\}}}$ denotes the group-relative advantage, which emphasizes outputs that surpass the group average for prioritized optimization.

\textbf{Reward Function Design.} We design reward functions to guide model learning from structured and accurate responses. For the \textbf{Format Reward}, we require the model to format its response using <think>...</think> for intermediate reasoning steps and <answer>...</answer> for the final answer. A reward score of 1 is assigned if both tag pairs appear exactly once and no content exists outside these tags; otherwise, the score is 0. For the \textbf{Accuracy Reward}, the content within the <answer>...</answer> tags is extracted using regular expressions,  and only exact matches to a valid option from the predefined set A, B, C, D, E, F are accepted. A reward score of 1 is assigned if the extracted answer matches the ground-truth label and 0 otherwise. For the \textbf{Length-Aware Penalty}, we implement a soft overlong punishment similar to the original work of DAPO. For more details on reward function design, see Appendix~\ref{appendix:lvlm training}.
For GRPO, the reward function is:

\begin{equation}
R^{\mathrm{GRPO}}(a_i) =
\begin{cases}
0.1 \cdot R_{\mathrm{fmt}}(a_i) + 0.9 \cdot R_{\mathrm{acc}}(a_i), & \text{if } R_{\mathrm{fmt}}(a_i) = 1 \land R_{\mathrm{acc}}(a_i) = 1 \\
0, & \text{otherwise}
\end{cases}
\end{equation}

Where $R_{\text{fmt}}$ denotes the format reward and $R_{\text{acc}}$ denotes the accuracy reward. For DAPO, the reward function is: 

\begin{equation}
R^{\mathrm{DAPO}}(a_i) =
\begin{cases}
0.5 \cdot R_{\mathrm{acc}}(a_i) + 0.5 \cdot R_{\mathrm{len}}(a_i), & \text{if } R_{\mathrm{acc}}(a_i) = 1 \land R_{\mathrm{len}}(a_i) = 1 \\ 
-1, & \text{otherwise}
\end{cases}
\end{equation}

Where $R_{\text{acc}}$ denotes the format reward and $R_{\text{len}}$ denotes the length-aware penalty.


\section{Experiments}

\subsection{Zero-shot cross-modal retrieval}

We compared PathoCLIP against ten baseline CLIP variants: OpenAI-CLIP-B, OpenAI-CLIP-L, PLIP, PathCLIP, CONCH, PathGen-CLIP, PathGen-CLIP-L, QuiltNet, PubmedCLIP, and MUSK. We evaluated each model on two image–caption pairing datasets: Archive, compiled in-house from diverse pathology texts, and ARCH \cite{arch}; performance was measured using Recall@K. As shown in Table~\ref{tab:retrieval_comparison}, on ARCH, PathoCLIP-L achieves the highest scores across all four Recall@K settings, obtaining mean i2t and t2i recalls of 62.28\% and 60.33\%, respectively, substantially outperforming the leading baseline, CONCH (50.71\% and 52.73\%); moreover, PathoCLIP-B also surpasses CONCH. In the Archive dataset, PathoCLIP-B attains mean i2t and t2i recalls of 13.34\% and 13.72\%, respectively, whereas PathoCLIP-L achieves 21.31\% and 22.13\%, far exceeding PubmedCLIP’s 9.18\% and 8.51\%. These results demonstrate that the latent-space alignment strategy of PathoCLIP confers state-of-the-art accuracy and robustness in cross-modal retrieval.

\begin{table}[ht]
  \centering
  \renewcommand{\arraystretch}{1.1}
  \setlength{\extrarowheight}{2pt}
  \captionsetup{justification=raggedright,singlelinecheck=false}
  \caption{Cross-modal retrieval results on the ARCH and Archive dataset. In each cell, results are displayed as “i2t / t2i” in the format (\%/\%). The best-performing results highlighted in bold, with the second-best underlined.}
  \resizebox{\textwidth}{!}{%
    \begin{tabular}{lcccccccccccc} 
      \toprule
      \multirow{2}{*}{\textbf{Model}} 
        & \multicolumn{6}{c}{\textbf{ARCH (@k)}}
        & \multicolumn{6}{c}{\textbf{Archive (@k)}} \\
      \cmidrule(lr){2-7} \cmidrule(lr){8-13}
        & \textbf{@1}       & \textbf{@5}        & \textbf{@10}       & \textbf{@20}       & \textbf{Aver(i2t)}       & \textbf{Aver(t2i)}
        & \textbf{@1}       & \textbf{@5}        & \textbf{@10}       & \textbf{@20}       & \textbf{Aver(i2t)}       & \textbf{Aver(t2i)} \\
      \midrule
      OpenAI-CLIP-B & 0.41/0.22 & 1.08/1.34 & 1.97/2.19 & 3.57/3.42 & 1.76 & 1.79 & 1.93/1.54 & 4.71/4.54 & 6.83/6.76 & 10.98/11.05 & 6.11 & 5.97 \\
      OpenAI-CLIP-L & 0.86/0.71 & 2.86/2.75 & 4.46/4.69 & 7.33/7.63 & 3.88 & 3.95 & 2.45/2.52 & 6.31/6.54 & 9.51/9.58 & 14.31/14.51 & 8.15 & 8.29 \\
      PLIP & 2.31/3.27 & 8.07/9.15 & 13.24/14.21 & 19.79/21.58 & 10.85 & 12.05 & 2.03/2.25 & 7.42/6.41 & 10.72/9.67 & 15.54/15.69 & 8.93 & 8.51 \\
      PathCLIP & 9.86/10.42 & 24.48/24.74 & 32.63/33.26 & 44.12/43.34 & 27.77 & 27.94 & 1.50/1.41 & 3.14/3.20 & 4.31/4.71 & 5.85/6.70 & 3.70 & 4.01 \\
      CONCH & 24.07/25.63 & 48.29/50.22 & 60.04/62.43 & 70.42/72.62 & 50.71 & 52.73 & 0.03/0.10 & 0.26/0.36 & 0.52/0.59 & 1.21/0.98 & 0.51 & 0.51 \\ 
      PathGen-CLIP & 14.17/14.99 & 34.15/32.29 & 45.2/41.44 & 56.99/51.75 & 37.63 & 35.12 & 0.62/0.62 & 1.70/1.67 & 2.81/2.58 & 3.69/4.18 & 2.21 & 2.26 \\
      PathGen-CLIP-L & 18.71/17.52 & 40.55/35.97 & 52.79/46.13 & 63.76/56.32 & 43.95 & 38.99 & 1.34/1.11 & 2.88/2.61 & 4.28/4.18 & 6.34/6.14 & 3.71 & 3.51 \\
      QuiltNet & 2.49/3.68 & 8.59/11.20 & 12.91/17.26 & 20.68/25.22 & 11.17 & 14.34 & 0.78/1.08 & 2.19/2.68 & 3.89/4.35 & 6.14/6.41 & 3.25 & 3.63 \\
      PubmedCLIP & 0.15/0.19 & 0.82/1.00 & 1.23/1.60 & 1.97/2.94 & 1.04 & 1.43 & 2.03/2.25 & 7.42/6.41 & 10.72/9.67 & 16.54/15.69 & 9.18 & 8.51 \\
      MUSK & 25.52/23.96 & 50.04/49.29 & 62.09/62.20 & 73.36/73.47 & 52.75 & 52.23 & 1.86/1.90 & 4.41/5.29 & 6.18/7.84 & 8.63/11.18 & 5.27 & 6.55 \\
      \rowcolor{purple!20}
      \textbf{Patho–CLIP–B}    & \underline{27.53}/\underline{26.71} & \underline{55.69}/\underline{55.13} & \underline{67.60}/\underline{67.08} & \underline{77.57}/\underline{77.08} & \underline{57.10} & \underline{56.50} & \underline{3.63}/\underline{3.43} & \underline{10.49}/\underline{10.78} & \underline{16.01}/\underline{16.50} & \underline{23.24}/\underline{24.15} & \underline{13.34} & \underline{13.72} \\
      \rowcolor{purple!20}
      \textbf{Patho–CLIP–L}    & \textbf{30.73/28.20} & \textbf{61.01/59.45} & \textbf{73.92/71.91} & \textbf{83.44/81.77} & \textbf{62.28} & \textbf{60.33}
                      & \textbf{6.27/6.14} & \textbf{17.29/18.27} & \textbf{25.62/26.93} & \textbf{36.05/37.19} & \textbf{21.31} & \textbf{22.13} \\
      \bottomrule
    \end{tabular}
  }
  \label{tab:retrieval_comparison}
\end{table}

\subsection{Zero-shot image classification}

To demonstrate PathoCLIP’s zero-shot classification performance, we evaluated it on five pathology datasets: SICAPv2 \cite{sicapv2}, WSSSLUAD \cite{wsss}, LC-Lung and LC-Colon \cite{lung}, and BMT \cite{bmt}. For each dataset, we augmented class labels with synonyms (e.g., in LC-Colon dataset, the “colon adenocarcinoma” class was enriched with terms such as “adenocarcinoma of the colon” and “colorectal adenocarcinoma”) and constructed category-specific text templates (e.g., “an H\&E image of CLASSNAME.”, “an image of CLASSNAME.”).

\begin{wraptable}{l}{\dimexpr0.6\linewidth\relax}
\vskip -0.5em
\caption{Comparison of different CLIP models on zero-shot pathology image classification datasets with accuracy (\%). The top performance is highlighted in bold, with the second-best underlined.}

\resizebox{\linewidth}{!}{%
    \begin{tabular}{lcccccc}
      \toprule
      \rowcolor{white}
      \textbf{Model}
        & \textbf{LC-Lung}
        & \textbf{LC-Colon}
        & \textbf{WSSSLUAD}
        & \textbf{SICAPv2}
        & \textbf{BMT}
        & \textbf{Average} \\
      \midrule
      OpenAI–CLIP–B   & 54.71 & 67.78 & 81.25 & 25.68 & 31.43 & 52.17 \\
      OpenAI–CLIP–L   & 70.34 & 70.25 & 76.25 & 15.41 & 26.27 & 51.70 \\
      PLIP            & 88.79 & 78.38 & 86.21 & 50.61 & 33.33 & 67.47 \\
      PathCLIP        & 86.74 & 96.29 & 91.25 & 39.30 & 34.33 & 69.58 \\
      CONCH           & 88.93 & 97.35 & 65.14 & 47.36 & 40.67 & 67.89 \\
      PathGen–CLIP    & 91.15 & 96.44 & 81.06 & \underline{54.15} & 37.17 & 71.99 \\
      PathGen–CLIP–L  & 91.85 & 98.30 & 82.50 & 53.39 & \underline{47.21} & 74.65 \\
      QuiltNet        & 62.73 & 58.15 & 90.50 & 41.85 & 33.83 & 57.41 \\
      PubmedCLIP      & 86.73 & 95.66 & 91.25 & 39.07 & 34.83 & 69.51 \\
      MUSK            & \underline{92.93} & \underline{98.54} & 87.50 & 44.96 & 45.33 & 73.25 \\
      \rowcolor{purple!20} 
      \textbf{Patho–CLIP–B}    & 89.90 & \textbf{98.60} & \underline{91.80} & 49.01 & \textbf{49.33} & \underline{75.28} \\
      \rowcolor{purple!20}
      \textbf{Patho–CLIP–L}
                      & \textbf{93.78}
                      & 97.48
                      & \textbf{95.89}
                      & \textbf{55.09}
                      & 38.44
                      & \textbf{76.14} \\
      \bottomrule
    \end{tabular}
    \label{table:2}
}
\end{wraptable}

 As shown in Table~\ref{table:2}, PathoCLIP-L surpasses CONCH by 30.75\% on WSSSLUAD dataset and by 7.73\% on SICAPv2 dataset, achieving the highest accuracy across all five datasets with a mean accuracy of 76.14\%. Notably, on BMT dataset, PathoCLIP-B attains 49.33\% accuracy—exceeding PathoCLIP-L and substantially outperforming QuiltNet (33.83\%) and PathGen-CLIP (37.17\%).  The strong performance of PathoCLIP underscores the diversity of our pathology training data and validates our two-stage progressive training paradigm for enhancing semantic understanding in CLIP models.

\subsection{Few-shot image classification with linear probing}

We also discuss the effectiveness of the model in few-shot learning, which is fine-tuned by linear probing of the pre-extracted feature representations of the model. We selected two representative datasets, LC-Lung and BMT, for testing under different training sizes of 2, 8, 16, 32, 64 and 128 samples, each of which implements 10 random samples. For each sample setting, we run 10 independent experiments, and finally the performance of the model under multivariate conditions is visualized by box-and-line plots.

\begin{figure}[ht]
    \centering
    \begin{subfigure}[b]{0.48\textwidth}
        \centering
        \includegraphics[width=\textwidth]{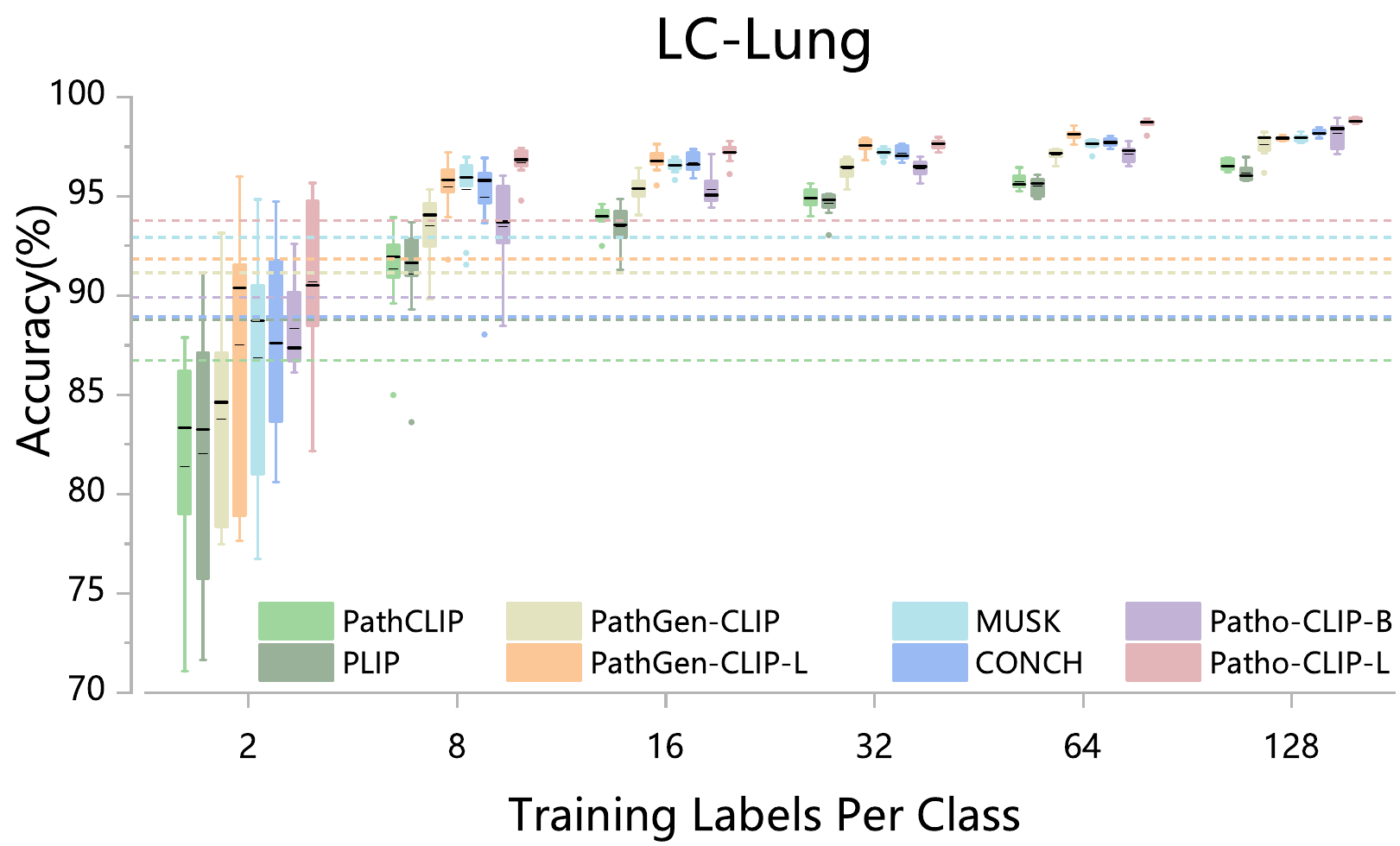}
        \label{fig:lc-lung}
    \end{subfigure}
    \hfill
    \begin{subfigure}[b]{0.48\textwidth}
        \centering
        \includegraphics[width=\textwidth]{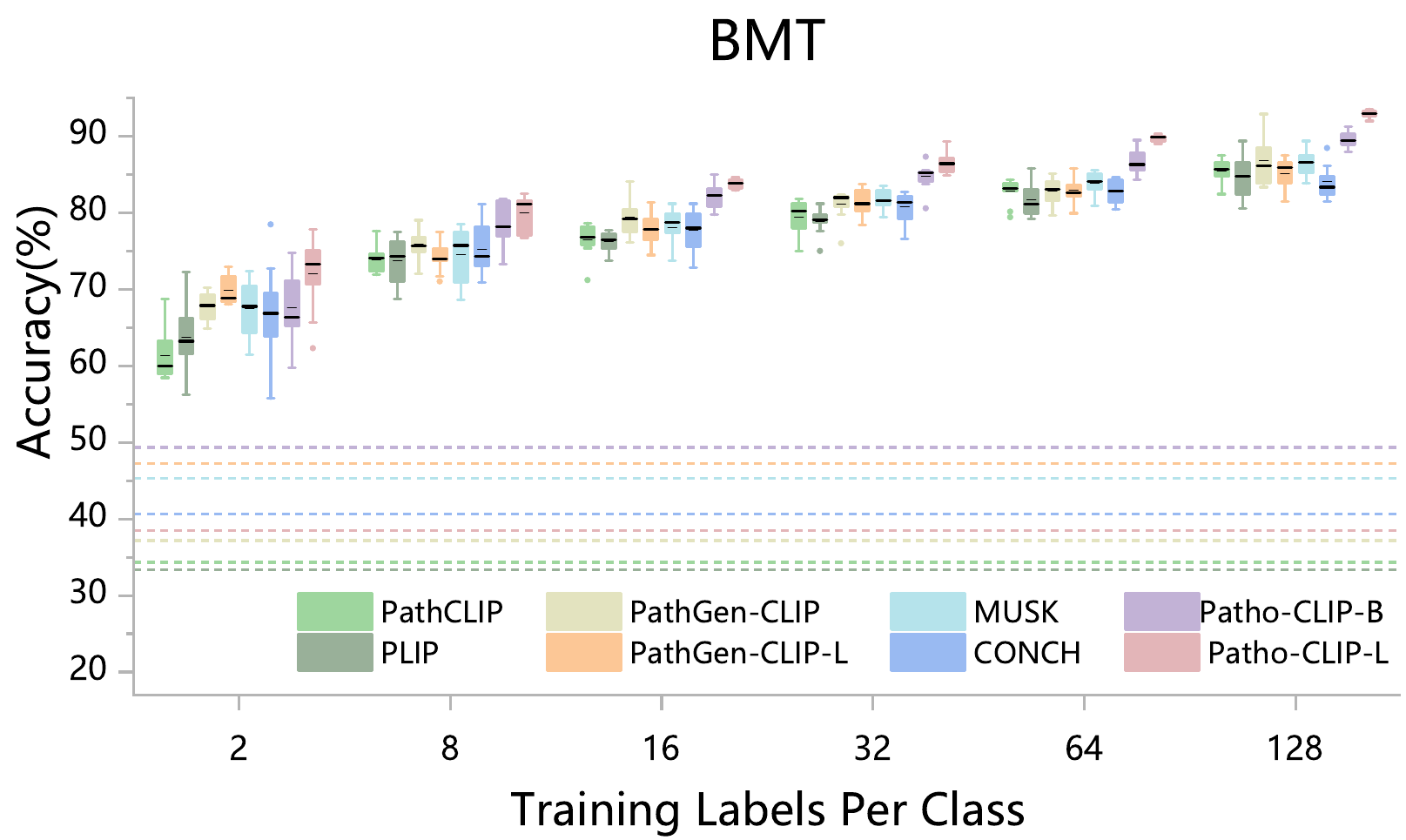}
        \label{fig:bmt}
    \end{subfigure}

    \caption{Comparison of few‐shot classification accuracy (\%) across different CLIP models on various pathology image classification datasets. For reference, the zero-shot performance of each model is shown as a dashed line on the same plot. }
    \label{fig:few_shot_pathology}
\end{figure}

The experimental results show that the PathoCLIP series demonstrates excellent performance in extreme low-sample scenarios, providing a viable solution for annotation-poor clinical environments. As shown in Figure~\ref{fig:few_shot_pathology}, on the BMT dataset, the PathoCLIP-L model achieves 73\% accuracy with only 2 training samples, which is a significant advantage over the comparative models such as CONCH, PLIP, MUSK, PathGen-CLIP, etc. (none of the comparative models exceeds 70\% accuracy), and the performance of the PathoCLIP-L model improves faster and converges faster as the number of training samples increases. The model outperforms comparable methods in terms of performance improvement and convergence speed as the number of training samples increases. The narrower distribution interval presented by its box-and-line plot further confirms the leading edge of the model in terms of task adaptability and robustness. Further experiments are provided in Appendix~\ref{appendix:clip evaluation}.

\subsection{Patho-R1 Evaluation Results}

\subsubsection{Open-ended Benchmarks}

\label{sec:Experiments}
\begin{figure}
    \centering
    \includegraphics[width=1\linewidth]{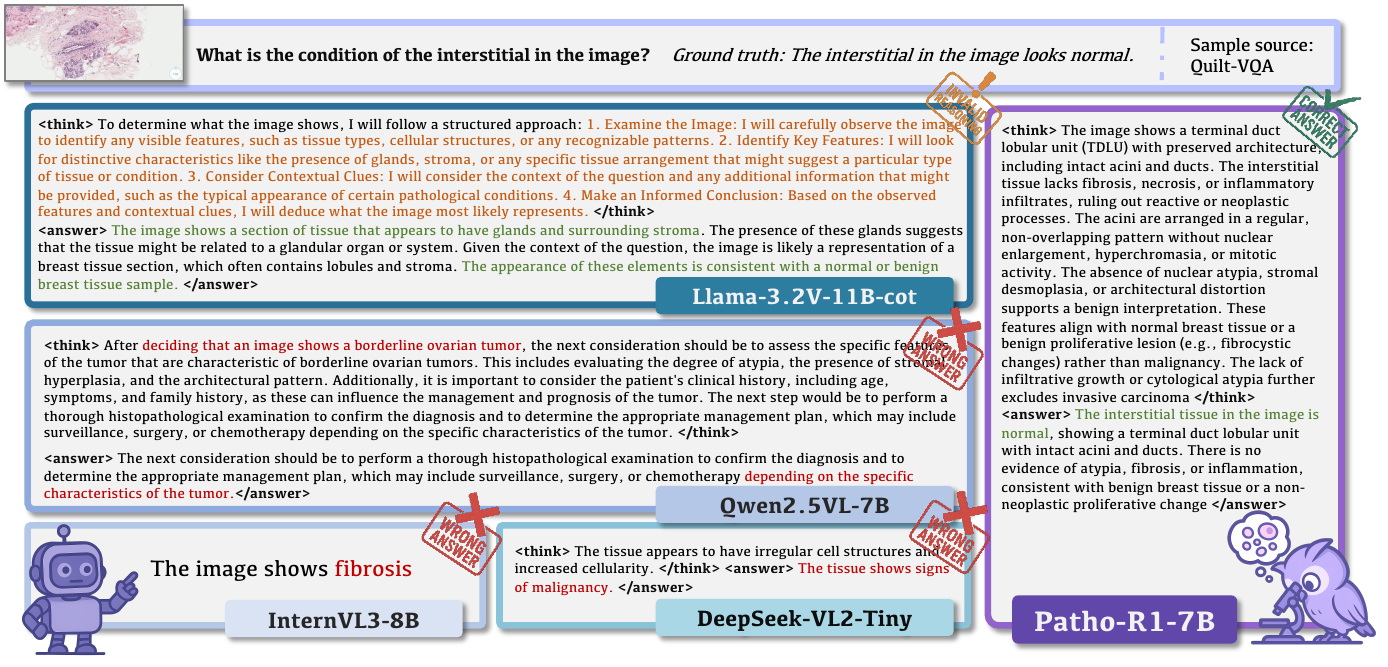}
    \caption{A representative example from the VQA task. While other models exhibit issues in answer formatting, factual accuracy, and reasoning coherence, our model consistently delivers well-structured, accurate, and logically sound responses. }
    \label{fig:vqa_example}
\end{figure}

Our proposed model, Patho-R1, is designed to perform step-by-step reasoning and produce accurate final answers. As shown in Figure~\ref{fig:vqa_example}, existing general-purpose and medical models do not necessarily possess the reasoning capability required to correctly answer pathology-specific questions. To evaluate its performance, we follow a fuzzy evaluation strategy in which Deepseek-R1 acts as the LLM-judge, scoring generated outputs based on ten well-defined criteria \cite{llamavo1}. Evaluations are conducted using the VLMEvalKit \cite{vlmevalkit} framework on two publicly available pathology VQA datasets: Quilt-VQA and Path-VQA. This setup ensures consistency, reproducibility, and fair comparison with other models. Our comprehensive evaluation highlights the strength of Patho-R1 in advancing multimodal reasoning in pathology. In addition to the traditional CoT prompts, we also used the latest Chain-of-Draft (CoD) prompts \cite{cod} to explore whether simplifying the reasoning chain affects model performance.

\begin{figure}[h]
    \centering
    \includegraphics[width=1\linewidth]{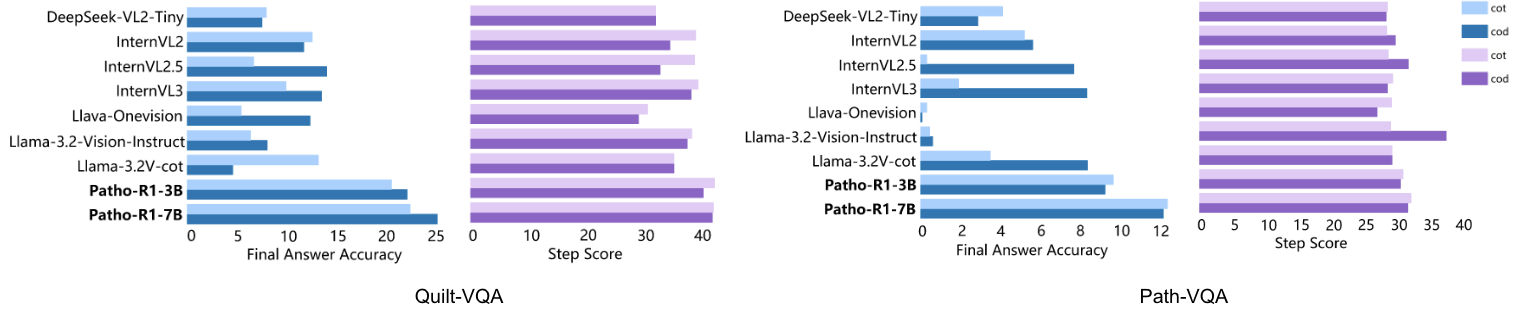}
    \caption{Evaluation of models with reasoning capabilities on final answer accuracy, coherence, and logical consistency of their reasoning steps using Quilt-VQA and Path-VQA.}
    \label{open-bench}
\end{figure}

Our experimental results in Figure~\ref{open-bench} show that Patho-R1 achieves the highest accuracy on both the Quilt VQA and Path VQA open-ended question-answering datasets, while also ranking highly in reasoning quality. An interesting finding is that CoD prompts do help improve answer accuracy, especially on the relatively simpler Quilt VQA dataset, where the overall accuracy is higher.

\subsubsection{Close-ended Benchmarks}

Closed-ended questions play a crucial role in pathology-related tasks, particularly in diagnostic classification. To evaluate model performance on such tasks, we consider two types of close-ended question datasets: (1) Yes/No questions, selected from Path-VQA and Quilt-VQA; and (2) multiple-choice questions, sourced from PathMMU \cite{pathmmu}, MedXpertQA \cite{medxpertqa}, and OmniMedVQA \cite{omnimedvqa}. For both types, we report accuracy as the evaluation metric to ensure consistency and comparability across datasets. Experimental results Table~\ref{mcq1}, Table~\ref{mcq2} demonstrate that Patho-R1 outperforms previous state-of-the-art pathology-focused LMMs. Specifically, it surpasses PathGen-LLaVA-13B by 7.63 \% on the PathMMU test-tiny split (PathGen-LLaVA-13B: 61.9\%), by approximately 5\% on the PathMMU test set and PathMMU evaluation set. For ablation studies and other experiments, see Appendix~\ref{appendix:multimodal benchmark for lvlms} and~\ref{appendix:ablation experiment}.

\begin{table}[htbp]
\caption{Comparison of different multimodels on PathMMU-test-tiny and PathMMU-test benchmarks. The top performance is highlighted in bold, with the second-best underlined.}
\resizebox{\textwidth}{!}{%
\begin{tabular}{l|rrrrrr|rrrrrr}
\toprule
\multirow{2}{*}{\textbf{Model}} & \multicolumn{6}{c|}{\textbf{PathMMU-test-tiny (1139)}} & \multicolumn{6}{c}{\textbf{PathMMU-test (8454)}} \\
\cmidrule(lr){2-7} \cmidrule(lr){8-13}
 & Atlas & EduContent & PathCLS & PubMed & SocialPath & Overall & Atlas & EduContent & PathCLS & PubMed & SocialPath & Overall \\
\midrule
\multicolumn{13}{l}{\textbf{Small model without thinking ability}} \\
Paligemma2-3b-pt-224 & 0.96 & 1.96 & 0 & 2.14 & 0.92 & 1.32 & 0.5 & 3.15 & 0.06 & 2.01 & 1.61 & 1.64 \\
VILA1.5-3B  & 28.85 & 29.41 & 16.38 & 34.88 & 23.85 & 27.57 & 26.53 & 30.60 & 15.07 & 29.10 & 30.71 & 26.74 \\
Llama3.2-3B-vl  & 30.29 & 27.84 & 7.34 & 27.76 & 30.28 & 25.55 & 35.79 & 25.01 & 6.99 & 25.37 & 28.91 & 23.39 \\
\midrule
\multicolumn{13}{l}{\textbf{Large model without thinking ability}} \\
LLaVA-Med-v1.5-Mistral-7B  & 25.00 & 21.57 & 7.34 & 24.56 & 22.48 & 20.9 & 21.40 & 21.51 & 7.72 & 21.31 & 20.54 & 18.59 \\
HuatuoGPT-Vision-7B  & 65.87 & 60.00 & 40.11 & 61.92 & 58.72 & 58.21 & 58.07 & 54.72 & 36.64 & \underline{61.36} & 59.37 & 54.59 \\
Quilt-LLaVA-v1.5-7B  & 42.79 & 38.43 & 14.12 & 37.01 & 32.57 & 33.98 & 41.43 & 36.72 & 14.71 & 34.80 & 35.29 & 32.02 \\
PathGen-LLaVA-13B  & 68.27 & 62.35 & \textbf{55.93} & 62.28 & 59.63 & 61.9 & 63.83 & 58.47 & \textbf{56.19} & 59.24 & 58.47 & \underline{58.79} \\
\midrule
\multicolumn{13}{l}{\textbf{Small model with thinking ability}} \\
DeepSeek-VL2-Tiny & 38.46 & 33.73 & 19.77 & 38.79 & 36.70 & 34.24 & 32.17 & 35.41 & 19.18 & 35.38 & 32.13 & 31.36 \\
Qwen2.5VL-3B & 46.63 & 44.31 & 23.73 & 49.82 & 46.79 & 43.37 & 45.06 & 45.28 & 23.65 & 46.79 & 48.10 & 42.10 \\
\rowcolor{purple!20}
\textbf{Patho-R1-3B} & \underline{74.52} & \underline{67.45} & 36.72 & \underline{67.62} & \underline{66.51} & \underline{63.83} & \underline{72.22} & \underline{62.51} & 34.99 & 61.07 & \underline{64.07} & 57.93 \\
\midrule
\multicolumn{13}{l}{\textbf{Large model with thinking ability}} \\
InternVL2-8B & 46.63 & 50.59 & 21.47 & 49.11 & 51.38 & 45.13 & 43.68 & 44.86 & 23.77 & 44.56 & 45.40 & 40.68 \\
InternVL2.5-8B & 51.44 & 50.59 & 29.38 & 55.87 & 57.80 & 50.13 & 50.06 & 50.62 & 32.84 & 50.02 & 50.87 & 46.98 \\
InternVL3-8B & 58.17 & 54.90 & 42.94 & 57.65 & 60.55 & 55.4 & 54.07 & 50.80 & 39.09 & 54.04 & 53.32 & 50.38 \\
Llama-3.2-11B-Vision-Instruct & 45.19 & 38.04 & 29.38 & 39.50 & 41.74 & 39.07 & 41.05 & 37.49 & 26.72 & 38.82 & 39.21 & 36.50 \\
Llama-3.2V-11B-cot & 49.04 & 47.06 & 29.94 & 53.38 & 45.41 & 46.01 & 51.81 & 45.45 & 30.76 & 48.15 & 46.10 & 44.23 \\
LLaVA-Onevision-7B & 31.25 & 21.18 & 13.56 & 31.32 & 18.35 & 23.79 & 21.65 & 21.27 & 12.01 & 27.77 & 21.25 & 23.40 \\
Qwen2.5VL-7B & 44.23 & 49.41 & 24.86 & 44.84 & 40.83 & 41.88 & 41.18 & 43.20 & 24.82 & 42.77 & 39.67 & 38.67 \\
\rowcolor{purple!20}
\textbf{Patho-R1-7B} & \textbf{81.73} & \textbf{75.29} & \underline{44.63} & \textbf{72.24} & \textbf{67.89} & \textbf{69.53} & \textbf{75.34} & \textbf{66.43} & \underline{45.40} & \textbf{66.06} & \textbf{67.93} & \textbf{63.37} \\
\bottomrule
\end{tabular}%
}
\label{mcq1}
\end{table}

\begin{table}[htbp]
\caption{Comparison of different multimodels on PathMMU-val, Quilt-VQA and Path-VQA benchmarks. The top performance is highlighted in bold, with the second-best underlined.}
\resizebox{\textwidth}{!}{%
\begin{tabular}{rlrrrrrrrr}
\toprule
\multicolumn{2}{c}{} & \multicolumn{6}{c}{\textbf{PathMMU-val (705)}} & \multicolumn{2}{c}{\textbf{YorN (3705)}} \\
\cmidrule(lr){3-8} \cmidrule(lr){9-10}
 & \textbf{Model} & Atlas & EduContent & PathCLS & PubMed & SocialPath & Overall & Quilt-VQA & Path-VQA \\
 \midrule
 \midrule
\multicolumn{10}{l}{\textbf{Small model without thinking ability}} \\
  & Paligemma2-3b-pt-224    & 1.25  & 1.37  & 0.00  & 2.15  & 0.67  & 1.28& 32.94 & 44.71 \\
  & VILA1.5-3B              & 22.50 & 27.40 & 16.67 & 31.33 & 30.00 & 27.23& 46.65 & 52.41 \\
  & Llama3.2-3B-vl          & 36.25 & 26.03 & 7.29  & 27.90 & 26.00 & 25.25& \textbf{69.39} & 55.44 \\
\midrule
\multicolumn{10}{l}{\textbf{Large model without thinking ability}} \\
  & LLaVA-Med-v1.5-Mistral-7B & 22.50 & 20.55 & 7.29  & 18.88 & 18.00 & 17.87& \underline{67.35} & 55.38 \\
  & HuatuoGPT-Vision-7B     & 61.25 & 54.11 & 38.54 & 55.36 & \underline{60.67} & 54.61& 59.18 & \textbf{65.85} \\
  & Quilt-LLaVA-v1.5-7B    & 45.00 & 32.88 & 17.71 & 34.33 & 34.67 & 33.05& 21.28 & 20.76 \\
  & PathGen-LLaVA-13B       & 67.50 & \underline{60.96} & \textbf{50.00} & 59.66 & 53.33 & 58.16& 47.52 & 45.75 \\
  
\multicolumn{10}{l}{\textbf{Small model with thinking ability}} \\
  & DeepSeek-VL2-Tiny          & 36.25 & 28.77 & 17.71 & 28.33 & 34.00 & 29.08& 44.31 & 46.34 \\
  & Qwen2.5VL-3B                    & 52.50 & 45.89 & 28.13 & 51.07 & 44.67 & 45.67& 40.52 & 43.72 \\
  \rowcolor{purple!20}
  & Patho-R1-3B                   & \underline{71.25} & 60.27 & 33.33 & \underline{62.66} & 59.33 & \underline{58.44}& 64.14 & 49.29 \\
\midrule
\multicolumn{10}{l}{\textbf{Large model with thinking ability}} \\
  & InternVL2-8B                 & 41.25 & 41.10 & 23.96 & 42.92 & 45.33 & 40.28& 63.56 & 61.36 \\
  & InternVL2.5-8B               & 46.25 & 52.05 & 33.33 & 47.64 & 54.00 & 47.80& 60.06 & \underline{64.78} \\
  & InternVL3-8B                  & 52.50 & 45.89 & 40.63 & 52.36 & 54.00 & 49.79& 33.82 & 18.56 \\
  & Llama-3.2-11B-Vision-Instruct & 46.25 & 36.30 & 32.29 & 36.05 & 39.33 & 37.45& 63.27 & 63.50 \\
  & Llama-3.2V-11B-cot            & 46.25 & 42.47 & 35.42 & 47.64 & 52.67 & 45.82& 54.81 & 56.42 \\
  & LLaVA-Onevision-7B            & 20.00 & 19.18 & 16.67 & 30.04 & 23.33 & 23.40& 24.20 & 52.38 \\
  & Qwen2.5VL-7B                   & 43.75 & 34.93 & 29.17 & 39.91 & 42.67 & 38.44& 52.19 & 41.82 \\
  \rowcolor{purple!20}
  & Patho-R1-7B                  & \textbf{82.50} & \textbf{63.01} & \underline{41.67} & \textbf{63.95} & \textbf{64.67} & \textbf{62.98}& 64.72 & 46.97 \\
\bottomrule
\end{tabular}%
}
\label{mcq2}
\end{table}

\section{Conclusion}
In this study, we propose a comprehensive data curation pipeline for the three phases of pathology VLM training. Using this pipeline, we construct an extensive corpus containing 3.5 million image-text pairs, 500k SFT data with reasoning, and 10k MCQs for RL. Furthermore, we explore the end-to-end domain adaptation of pretrained VLMs, yielding Patho-R1 and Patho-CLIP, two open-source models that contribute to the advancement of pathology-specific vision-language research and offer practical insights for future development in this domain.

\textbf{Limitations.} First, while our CPT method effectively enhances pathology-specific alignment, more advanced CPT strategies (e.g., \cite{cheng2024adaptinglargelanguagemodels, huatuogptvision}) could further mitigate instruction-following degradation but were not adopted due to the high computational cost of processing large corpora. Second, although our models exhibit some generalization ability, their performance degrades on out-of-domain modalities (e.g., MRI or CT), as our training data focuses exclusively on pathology-related sources.
\label{sec:limitation}


{\small
\bibliographystyle{IEEEtran}
\bibliography{pathreason}

\begin{thebibliography}{10}
\providecommand{\url}[1]{#1}
\csname url@samestyle\endcsname
\providecommand{\newblock}{\relax}
\providecommand{\bibinfo}[2]{#2}
\providecommand{\BIBentrySTDinterwordspacing}{\spaceskip=0pt\relax}
\providecommand{\BIBentryALTinterwordstretchfactor}{4}
\providecommand{\BIBentryALTinterwordspacing}{\spaceskip=\fontdimen2\font plus
\BIBentryALTinterwordstretchfactor\fontdimen3\font minus \fontdimen4\font\relax}
\providecommand{\BIBforeignlanguage}[2]{{%
\expandafter\ifx\csname l@#1\endcsname\relax
\typeout{** WARNING: IEEEtran.bst: No hyphenation pattern has been}%
\typeout{** loaded for the language `#1'. Using the pattern for}%
\typeout{** the default language instead.}%
\else
\language=\csname l@#1\endcsname
\fi
#2}}
\providecommand{\BIBdecl}{\relax}
\BIBdecl

\bibitem{tolkach2023artificial}
Y.~Tolkach, L.~M. Wolgast, A.~Damanakis, A.~Pryalukhin, S.~Schallenberg, W.~Hulla, M.-L. Eich, W.~Schroeder, A.~Mukhopadhyay, M.~Fuchs \emph{et~al.}, ``Artificial intelligence for tumour tissue detection and histological regression grading in oesophageal adenocarcinomas: a retrospective algorithm development and validation study,'' \emph{The Lancet Digital Health}, vol.~5, no.~5, pp. e265--e275, 2023.

\bibitem{foersch2023multistain}
S.~Foersch, C.~Glasner, A.-C. Woerl, M.~Eckstein, D.-C. Wagner, S.~Schulz, F.~Kellers, A.~Fernandez, K.~Tserea, M.~Kloth \emph{et~al.}, ``Multistain deep learning for prediction of prognosis and therapy response in colorectal cancer,'' \emph{Nature medicine}, vol.~29, no.~2, pp. 430--439, 2023.

\bibitem{wang2025triad}
S.~Wang, M.~Safari, Q.~Li, C.-W. Chang, R.~L. Qiu, J.~Roper, D.~S. Yu, and X.~Yang, ``Triad: Vision foundation model for 3d magnetic resonance imaging,'' \emph{arXiv preprint arXiv:2502.14064}, 2025.

\bibitem{wu2024voco}
L.~Wu, J.~Zhuang, and H.~Chen, ``Voco: A simple-yet-effective volume contrastive learning framework for 3d medical image analysis,'' in \emph{Proceedings of the IEEE/CVF Conference on Computer Vision and Pattern Recognition}, 2024, pp. 22\,873--22\,882.

\bibitem{xie2024medtrinity}
Y.~Xie, C.~Zhou, L.~Gao, J.~Wu, X.~Li, H.-Y. Zhou, S.~Liu, L.~Xing, J.~Zou, C.~Xie \emph{et~al.}, ``Medtrinity-25m: A large-scale multimodal dataset with multigranular annotations for medicine,'' \emph{arXiv preprint arXiv:2408.02900}, 2024.

\bibitem{CLIP}
A.~Radford, J.~W. Kim, C.~Hallacy, A.~Ramesh, G.~Goh, S.~Agarwal, G.~Sastry, A.~Askell, P.~Mishkin, J.~Clark \emph{et~al.}, ``Learning transferable visual models from natural language supervision,'' in \emph{International conference on machine learning}.\hskip 1em plus 0.5em minus 0.4em\relax PmLR, 2021, pp. 8748--8763.

\bibitem{LLAVA}
H.~Liu, C.~Li, Q.~Wu, and Y.~J. Lee, ``Visual instruction tuning,'' \emph{Advances in neural information processing systems}, vol.~36, pp. 34\,892--34\,916, 2023.

\bibitem{medclip}
Z.~Wang, Z.~Wu, D.~Agarwal, and J.~Sun, ``Medclip: Contrastive learning from unpaired medical images and text,'' in \emph{Proceedings of the Conference on Empirical Methods in Natural Language Processing. Conference on Empirical Methods in Natural Language Processing}, vol. 2022, 2022, p. 3876.

\bibitem{medflamingo}
M.~Moor, Q.~Huang, S.~Wu, M.~Yasunaga, Y.~Dalmia, J.~Leskovec, C.~Zakka, E.~P. Reis, and P.~Rajpurkar, ``Med-flamingo: a multimodal medical few-shot learner,'' in \emph{Machine Learning for Health (ML4H)}.\hskip 1em plus 0.5em minus 0.4em\relax PMLR, 2023, pp. 353--367.

\bibitem{radfm}
C.~Wu, X.~Zhang, Y.~Zhang, Y.~Wang, and W.~Xie, ``Towards generalist foundation model for radiology by leveraging web-scale 2d\&3d medical data,'' \emph{arXiv preprint arXiv:2308.02463}, 2023.

\bibitem{pmcclip}
W.~Lin, Z.~Zhao, X.~Zhang, C.~Wu, Y.~Zhang, Y.~Wang, and W.~Xie, ``Pmc-clip: Contrastive language-image pre-training using biomedical documents,'' in \emph{International Conference on Medical Image Computing and Computer-Assisted Intervention}.\hskip 1em plus 0.5em minus 0.4em\relax Springer, 2023, pp. 525--536.

\bibitem{biomedgpt}
K.~Zhang, R.~Zhou, E.~Adhikarla, Z.~Yan, Y.~Liu, J.~Yu, Z.~Liu, X.~Chen, B.~D. Davison, H.~Ren \emph{et~al.}, ``A generalist vision--language foundation model for diverse biomedical tasks,'' \emph{Nature Medicine}, pp. 1--13, 2024.

\bibitem{llavamed}
C.~Li, C.~Wong, S.~Zhang, N.~Usuyama, H.~Liu, J.~Yang, T.~Naumann, H.~Poon, and J.~Gao, ``Llava-med: Training a large language-and-vision assistant for biomedicine in one day,'' \emph{Advances in Neural Information Processing Systems}, vol.~36, pp. 28\,541--28\,564, 2023.

\bibitem{quiltllava}
M.~S. Seyfioglu, W.~O. Ikezogwo, F.~Ghezloo, R.~Krishna, and L.~Shapiro, ``Quilt-llava: Visual instruction tuning by extracting localized narratives from open-source histopathology videos,'' in \emph{Proceedings of the IEEE/CVF Conference on Computer Vision and Pattern Recognition}, 2024, pp. 13\,183--13\,192.

\bibitem{pathgen16m}
\BIBentryALTinterwordspacing
Y.~Sun, Y.~Zhang, Y.~Si, C.~Zhu, K.~Zhang, Z.~Shui, J.~Li, X.~Gong, X.~LYU, T.~Lin, and L.~Yang, ``Pathgen-1.6m: 1.6 million pathology image-text pairs generation through multi-agent collaboration,'' in \emph{The Thirteenth International Conference on Learning Representations}, 2025. [Online]. Available: \url{https://openreview.net/forum?id=rFpZnn11gj}
\BIBentrySTDinterwordspacing

\bibitem{quilt1m}
W.~Ikezogwo, S.~Seyfioglu, F.~Ghezloo, D.~Geva, F.~Sheikh~Mohammed, P.~K. Anand, R.~Krishna, and L.~Shapiro, ``Quilt-1m: One million image-text pairs for histopathology,'' \emph{Advances in neural information processing systems}, vol.~36, pp. 37\,995--38\,017, 2023.

\bibitem{plip}
Z.~Huang, F.~Bianchi, M.~Yuksekgonul, T.~J. Montine, and J.~Zou, ``A visual--language foundation model for pathology image analysis using medical twitter,'' \emph{Nature medicine}, vol.~29, no.~9, pp. 2307--2316, 2023.

\bibitem{gunasekar2023textbooksneed}
\BIBentryALTinterwordspacing
S.~Gunasekar, Y.~Zhang, J.~Aneja, C.~C.~T. Mendes, A.~D. Giorno, S.~Gopi, M.~Javaheripi, P.~Kauffmann, G.~de~Rosa, O.~Saarikivi, A.~Salim, S.~Shah, H.~S. Behl, X.~Wang, S.~Bubeck, R.~Eldan, A.~T. Kalai, Y.~T. Lee, and Y.~Li, ``Textbooks are all you need,'' 2023. [Online]. Available: \url{https://arxiv.org/abs/2306.11644}
\BIBentrySTDinterwordspacing

\bibitem{deepseekr1}
D.~Guo, D.~Yang, H.~Zhang, J.~Song, R.~Zhang, R.~Xu, Q.~Zhu, S.~Ma, P.~Wang, X.~Bi \emph{et~al.}, ``Deepseek-r1: Incentivizing reasoning capability in llms via reinforcement learning,'' \emph{arXiv preprint arXiv:2501.12948}, 2025.

\bibitem{qwq32b}
\BIBentryALTinterwordspacing
Q.~Team, ``Qwq-32b: Embracing the power of reinforcement learning,'' March 2025. [Online]. Available: \url{https://qwenlm.github.io/blog/qwq-32b/}
\BIBentrySTDinterwordspacing

\bibitem{kimik15}
K.~Team, A.~Du, B.~Gao, B.~Xing, C.~Jiang, C.~Chen, C.~Li, C.~Xiao, C.~Du, C.~Liao \emph{et~al.}, ``Kimi k1. 5: Scaling reinforcement learning with llms,'' \emph{arXiv preprint arXiv:2501.12599}, 2025.

\bibitem{visionr1incentivizing}
W.~Huang, B.~Jia, Z.~Zhai, S.~Cao, Z.~Ye, F.~Zhao, Z.~Xu, Y.~Hu, and S.~Lin, ``Vision-r1: Incentivizing reasoning capability in multimodal large language models,'' \emph{arXiv preprint arXiv:2503.06749}, 2025.

\bibitem{medr1}
Y.~Lai, J.~Zhong, M.~Li, S.~Zhao, and X.~Yang, ``Med-r1: Reinforcement learning for generalizable medical reasoning in vision-language models,'' \emph{arXiv preprint arXiv:2503.13939}, 2025.

\bibitem{medvlmr1}
J.~Pan, C.~Liu, J.~Wu, F.~Liu, J.~Zhu, H.~B. Li, C.~Chen, C.~Ouyang, and D.~Rueckert, ``Medvlm-r1: Incentivizing medical reasoning capability of vision-language models (vlms) via reinforcement learning,'' \emph{arXiv preprint arXiv:2502.19634}, 2025.

\bibitem{finr1}
\BIBentryALTinterwordspacing
Z.~Liu, X.~Guo, F.~Lou, L.~Zeng, J.~Niu, Z.~Wang, J.~Xu, W.~Cai, Z.~Yang, X.~Zhao, C.~Li, S.~Xu, D.~Chen, Y.~Chen, Z.~Bai, and L.~Zhang, ``Fin-r1: A large language model for financial reasoning through reinforcement learning,'' 2025. [Online]. Available: \url{https://arxiv.org/abs/2503.16252}
\BIBentrySTDinterwordspacing

\bibitem{Dapo}
Q.~Yu, Z.~Zhang, R.~Zhu, Y.~Yuan, X.~Zuo, Y.~Yue, T.~Fan, G.~Liu, L.~Liu, X.~Liu \emph{et~al.}, ``Dapo: An open-source llm reinforcement learning system at scale,'' \emph{arXiv preprint arXiv:2503.14476}, 2025.

\bibitem{pubmedclip}
S.~Eslami, C.~Meinel, and G.~De~Melo, ``Pubmedclip: How much does clip benefit visual question answering in the medical domain?'' in \emph{Findings of the Association for Computational Linguistics: EACL 2023}, 2023, pp. 1181--1193.

\bibitem{BiomedCLIP}
S.~Zhang, Y.~Xu, N.~Usuyama, H.~Xu, J.~Bagga, R.~Tinn, S.~Preston, R.~Rao, M.~Wei, N.~Valluri \emph{et~al.}, ``A multimodal biomedical foundation model trained from fifteen million image--text pairs,'' \emph{NEJM AI}, vol.~2, no.~1, p. AIoa2400640, 2025.

\bibitem{unimedclip}
M.~U. Khattak, S.~Kunhimon, M.~Naseer, S.~Khan, and F.~S. Khan, ``Unimed-clip: Towards a unified image-text pretraining paradigm for diverse medical imaging modalities,'' \emph{arXiv preprint arXiv:2412.10372}, 2024.

\bibitem{medpalm}
K.~Singhal, S.~Azizi, T.~Tu, S.~S. Mahdavi, J.~Wei, H.~W. Chung, N.~Scales, A.~Tanwani, H.~Cole-Lewis, S.~Pfohl \emph{et~al.}, ``Large language models encode clinical knowledge,'' \emph{Nature}, vol. 620, no. 7972, pp. 172--180, 2023.

\bibitem{huatuogptvision}
J.~Chen, C.~Gui, R.~Ouyang, A.~Gao, S.~Chen, G.~Chen, X.~Wang, Z.~Cai, K.~Ji, X.~Wan \emph{et~al.}, ``Towards injecting medical visual knowledge into multimodal llms at scale,'' in \emph{Proceedings of the 2024 Conference on Empirical Methods in Natural Language Processing}, 2024, pp. 7346--7370.

\bibitem{conch}
M.~Y. Lu, B.~Chen, D.~F. Williamson, R.~J. Chen, I.~Liang, T.~Ding, G.~Jaume, I.~Odintsov, L.~P. Le, G.~Gerber \emph{et~al.}, ``A visual-language foundation model for computational pathology,'' \emph{Nature Medicine}, vol.~30, no.~3, pp. 863--874, 2024.

\bibitem{musk}
J.~Xiang, X.~Wang, X.~Zhang, Y.~Xi, F.~Eweje, Y.~Chen, Y.~Li, C.~Bergstrom, M.~Gopaulchan, T.~Kim \emph{et~al.}, ``A vision--language foundation model for precision oncology,'' \emph{Nature}, pp. 1--10, 2025.

\bibitem{titan}
T.~Ding, S.~J. Wagner, A.~H. Song, R.~J. Chen, M.~Y. Lu, A.~Zhang, A.~J. Vaidya, G.~Jaume, M.~Shaban, A.~Kim \emph{et~al.}, ``Multimodal whole slide foundation model for pathology,'' \emph{arXiv preprint arXiv:2411.19666}, 2024.

\bibitem{pathasst}
Y.~Sun, C.~Zhu, S.~Zheng, K.~Zhang, L.~Sun, Z.~Shui, Y.~Zhang, H.~Li, and L.~Yang, ``Pathasst: A generative foundation ai assistant towards artificial general intelligence of pathology,'' in \emph{Proceedings of the AAAI Conference on Artificial Intelligence}, vol.~38, no.~5, 2024, pp. 5034--5042.

\bibitem{pathchat}
M.~Y. Lu, B.~Chen, D.~F. Williamson, R.~J. Chen, M.~Zhao, A.~K. Chow, K.~Ikemura, A.~Kim, D.~Pouli, A.~Patel \emph{et~al.}, ``A multimodal generative ai copilot for human pathology,'' \emph{Nature}, vol. 634, no. 8033, pp. 466--473, 2024.

\bibitem{cpathomni}
Y.~Sun, Y.~Si, C.~Zhu, X.~Gong, K.~Zhang, P.~Chen, Y.~Zhang, Z.~Shui, T.~Lin, and L.~Yang, ``Cpath-omni: A unified multimodal foundation model for patch and whole slide image analysis in computational pathology,'' \emph{arXiv preprint arXiv:2412.12077}, 2024.

\bibitem{pallava}
D.~Dai, Y.~Zhang, L.~Xu, Q.~Yang, X.~Shen, S.~Xia, and G.~Wang, ``Pa-llava: A large language-vision assistant for human pathology image understanding,'' in \emph{2024 IEEE International Conference on Bioinformatics and Biomedicine (BIBM)}.\hskip 1em plus 0.5em minus 0.4em\relax IEEE, 2024, pp. 3138--3143.

\bibitem{openaio1}
A.~Jaech, A.~Kalai, A.~Lerer, A.~Richardson, A.~El-Kishky, A.~Low, A.~Helyar, A.~Madry, A.~Beutel, A.~Carney \emph{et~al.}, ``Openai o1 system card,'' \emph{arXiv preprint arXiv:2412.16720}, 2024.

\bibitem{llavacot}
\BIBentryALTinterwordspacing
G.~Xu, P.~Jin, H.~Li, Y.~Song, L.~Sun, and L.~Yuan, ``Llava-cot: Let vision language models reason step-by-step,'' 2024. [Online]. Available: \url{https://arxiv.org/abs/2411.10440}
\BIBentrySTDinterwordspacing

\bibitem{llamavo1}
O.~Thawakar, D.~Dissanayake, K.~More, R.~Thawkar, A.~Heakl, N.~Ahsan, Y.~Li, M.~Zumri, J.~Lahoud, R.~M. Anwer \emph{et~al.}, ``Llamav-o1: Rethinking step-by-step visual reasoning in llms,'' \emph{arXiv preprint arXiv:2501.06186}, 2025.

\bibitem{visualrft}
Z.~Liu, Z.~Sun, Y.~Zang, X.~Dong, Y.~Cao, H.~Duan, D.~Lin, and J.~Wang, ``Visual-rft: Visual reinforcement fine-tuning,'' \emph{arXiv preprint arXiv:2503.01785}, 2025.

\bibitem{vlmr1}
H.~Shen, P.~Liu, J.~Li, C.~Fang, Y.~Ma, J.~Liao, Q.~Shen, Z.~Zhang, K.~Zhao, Q.~Zhang \emph{et~al.}, ``Vlm-r1: A stable and generalizable r1-style large vision-language model,'' \emph{arXiv preprint arXiv:2504.07615}, 2025.

\bibitem{perceptionr1}
E.~Yu, K.~Lin, L.~Zhao, J.~Yin, Y.~Wei, Y.~Peng, H.~Wei, J.~Sun, C.~Han, Z.~Ge \emph{et~al.}, ``Perception-r1: Pioneering perception policy with reinforcement learning,'' \emph{arXiv preprint arXiv:2504.07954}, 2025.

\bibitem{r1onevision}
Y.~Yang, X.~He, H.~Pan, X.~Jiang, Y.~Deng, X.~Yang, H.~Lu, D.~Yin, F.~Rao, M.~Zhu \emph{et~al.}, ``R1-onevision: Advancing generalized multimodal reasoning through cross-modal formalization,'' \emph{arXiv preprint arXiv:2503.10615}, 2025.

\bibitem{r1v}
L.~Chen, L.~Li, H.~Zhao, Y.~Song, and Vinci, ``R1-v: Reinforcing super generalization ability in vision-language models with less than \$3,'' \url{https://github.com/Deep-Agent/R1-V}, 2025, accessed: 2025-02-02.

\bibitem{zhao2024doclayoutyoloenhancingdocumentlayout}
\BIBentryALTinterwordspacing
Z.~Zhao, H.~Kang, B.~Wang, and C.~He, ``Doclayout-yolo: Enhancing document layout analysis through diverse synthetic data and global-to-local adaptive perception,'' 2024. [Online]. Available: \url{https://arxiv.org/abs/2410.12628}
\BIBentrySTDinterwordspacing

\bibitem{parmar2024reuse}
J.~Parmar, S.~Satheesh, M.~Patwary, M.~Shoeybi, and B.~Catanzaro, ``Reuse, don't retrain: A recipe for continued pretraining of language models,'' \emph{arXiv preprint arXiv:2407.07263}, 2024.

\bibitem{cheng2024adaptinglargelanguagemodels}
\BIBentryALTinterwordspacing
D.~Cheng, S.~Huang, and F.~Wei, ``Adapting large language models to domains via reading comprehension,'' 2024. [Online]. Available: \url{https://arxiv.org/abs/2309.09530}
\BIBentrySTDinterwordspacing

\bibitem{efficientnet}
\BIBentryALTinterwordspacing
M.~Tan and Q.~Le, ``{E}fficient{N}et: Rethinking model scaling for convolutional neural networks,'' in \emph{Proceedings of the 36th International Conference on Machine Learning}, ser. Proceedings of Machine Learning Research, K.~Chaudhuri and R.~Salakhutdinov, Eds., vol.~97.\hskip 1em plus 0.5em minus 0.4em\relax PMLR, 09--15 Jun 2019, pp. 6105--6114. [Online]. Available: \url{https://proceedings.mlr.press/v97/tan19a.html}
\BIBentrySTDinterwordspacing

\bibitem{autocluster}
\BIBentryALTinterwordspacing
Z.~Zhang, A.~Zhang, M.~Li, and A.~Smola, ``Automatic chain of thought prompting in large language models,'' in \emph{The Eleventh International Conference on Learning Representations}, 2023. [Online]. Available: \url{https://openreview.net/forum?id=5NTt8GFjUHkr}
\BIBentrySTDinterwordspacing

\bibitem{arch}
J.~Gamper and N.~Rajpoot, ``Multiple instance captioning: Learning representations from histopathology textbooks and articles,'' in \emph{Proceedings of the IEEE/CVF Conference on Computer Vision and Pattern Recognition (CVPR)}, June 2021, pp. 16\,549--16\,559.

\bibitem{sicapv2}
J.~Silva-Rodr{\'\i}guez, ``Sicapv2-prostate whole slide images with gleason grades annotations,'' \emph{Mendeley Data}, 2020.

\bibitem{wsss}
C.~Han, X.~Pan, L.~Yan, H.~Lin, B.~Li, S.~Yao, S.~Lv, Z.~Shi, J.~Mai, J.~Lin \emph{et~al.}, ``Wsss4luad: Grand challenge on weakly-supervised tissue semantic segmentation for lung adenocarcinoma,'' \emph{arXiv preprint arXiv:2204.06455}, 2022.

\bibitem{lung}
A.~A. Borkowski, M.~M. Bui, L.~B. Thomas, C.~P. Wilson, L.~A. DeLand, and S.~M. Mastorides, ``Lung and colon cancer histopathological image dataset (lc25000),'' \emph{arXiv preprint arXiv:1912.12142}, 2019.

\bibitem{bmt}
E.~C. Welch, C.~Lu, C.~J. Sung, C.~Zhang, A.~Tripathi, and J.~Ou, ``Bmt: A cross-validated thinprep pap cervical cytology dataset for machine learning model training and validation,'' \emph{Scientific Data}, vol.~11, no.~1, p. 1444, 2024.

\bibitem{vlmevalkit}
H.~Duan, J.~Yang, Y.~Qiao, X.~Fang, L.~Chen, Y.~Liu, X.~Dong, Y.~Zang, P.~Zhang, J.~Wang \emph{et~al.}, ``Vlmevalkit: An open-source toolkit for evaluating large multi-modality models,'' in \emph{Proceedings of the 32nd ACM International Conference on Multimedia}, 2024, pp. 11\,198--11\,201.

\bibitem{cod}
S.~Xu, W.~Xie, L.~Zhao, and P.~He, ``Chain of draft: Thinking faster by writing less,'' \emph{arXiv preprint arXiv:2502.18600}, 2025.

\bibitem{pathmmu}
Y.~Sun, H.~Wu, C.~Zhu, S.~Zheng, Q.~Chen, K.~Zhang, Y.~Zhang, D.~Wan, X.~Lan, M.~Zheng \emph{et~al.}, ``Pathmmu: A massive multimodal expert-level benchmark for understanding and reasoning in pathology,'' in \emph{European Conference on Computer Vision}.\hskip 1em plus 0.5em minus 0.4em\relax Springer, 2024, pp. 56--73.

\bibitem{medxpertqa}
Y.~Zuo, S.~Qu, Y.~Li, Z.~Chen, X.~Zhu, E.~Hua, K.~Zhang, N.~Ding, and B.~Zhou, ``Medxpertqa: Benchmarking expert-level medical reasoning and understanding,'' \emph{arXiv preprint arXiv:2501.18362}, 2025.

\bibitem{omnimedvqa}
Y.~Hu, T.~Li, Q.~Lu, W.~Shao, J.~He, Y.~Qiao, and P.~Luo, ``Omnimedvqa: A new large-scale comprehensive evaluation benchmark for medical lvlm,'' in \emph{Proceedings of the IEEE/CVF Conference on Computer Vision and Pattern Recognition}, 2024, pp. 22\,170--22\,183.

\end{thebibliography}
}

\newpage



\appendix


\newpage
\section{Datasets Construction} \label{appendix:Datasets construction}

In Appendix~\ref{sec:a1}, we detail the data cleaning strategies and classification schemes employed during training. Appendix~\ref{appendix:image-text pair sources} describes all image-text pairs used in the Continued Pretrain (CPT) stage, including both the textbook-derived datasets—along with the extraction methodology—and relevant publicly available datasets. Appendix~\ref{appendix:cot-sft data generation scheme} details the construction and distributional characteristics of the Supervised Fine-tuning (SFT) dataset, including the categorization heuristics, prompt design, and chain-of-thought (CoT) annotation strategies, and some examples. Appendix~\ref{appendix:rl dataset construction} details the data generation and sampling distributions for the RL stage. 

\subsection{Datasets Quality Control} \label{sec:a1}
\subsubsection{Data Cleaning}

During the data preprocessing stage, we encountered several quality issues that required targeted cleaning strategies to ensure the integrity of the dataset used for both supervised fine-tuning and reinforcement learning (RL).

\textbf{Repetitive Token Generation:} A common issue was pathological repetition, where the model generated outputs with excessive consecutive repetition of specific non-functional tokens (e.g., "cell", "nucleus"). To detect such cases, we analyzed each generated sample for consecutively repeated words or phrases. If any content-bearing token appeared more than 15 times in direct succession, the output was flagged for manual review. Flagged samples were then examined and filtered to remove degenerate outputs.

\textbf{Code-Switching and Translation Errors:} In the OCR and translation pipeline, the LLaMA-3-7B model, which we used for translation, occasionally failed to translate rare Chinese pathological terminologies into English, resulting in mixed-language outputs. To mitigate this, we implemented a rule-based character-level filter to detect residual Chinese characters in the translated text. Detected samples were reprocessed using a more capable model (Qwen-Max) iteratively, until all Chinese content was successfully translated.

\textbf{Misleading Visual Referencing in Captions:} When generating prompts for visual instruction tuning, we observed that the text-only language model tended to refer to image captions as “the description” rather than acknowledging them as visual content. This mismatch in modality reference could lead to incorrect downstream model behavior. To resolve this, we explicitly modified the prompt template to instruct the model to refer to all visual descriptions as being “in the image” rather than “in the description,” reinforcing the grounding to visual input. For residual cases where such wrong references still occurred, we implemented scanned model outputs for keywords such as `description', `mention', and systematically replaced them with the term `image' when appropriate. 

\textbf{Redundant Sampling Due to Caption Prefix Duplication:} In RL data sampling for multi-panel figures, a shared textual prefix is often attached to each sub-caption (e.g., preceding panel labels such as (a), (b)). Ideally, sub-captions retain distinct information following the prefix. However, when the unique content is short and the shared prefix is long, many captions become nearly indistinguishable, leading to redundancy. To mitigate this, we identified and removed samples whose first 50 characters were identical, preserving only one representative per group.

\textbf{Repetitive Question Generation During RL Sampling:} During the RL phase, we employed batched API calls to generate MCQ data using the DeepSeek-V3. Although each prompt included an example list of possible questions, we observed that the model occasionally produced identical questions across consecutive dialogues within the same batch. To address this, we curated multiple semantically equivalent but lexically diverse versions of the question list. These alternative phrasings were manually injected to replace repeated instances, thereby enhancing question variety and improving the robustness of the training data.

\subsubsection{Pathology Subfield Classification} \label{sec:a12}

The image-text pairs extracted from textbooks are categorized into the following 11 subfields (10 auto-classified by the classifier and 1 inspected manually):
(1) Histopathology: Hematoxylin and eosin stained (H\&E) microscopic images illustrating tissue morphology and cellular details for pathological diagnosis.
(2) Gross examination: Macroscopic images of surgical specimens, showcasing anatomical structures, lesion size, and surface characteristics.
(3) Cytology: Microscopic images of individual or clustered cells obtained through techniques such as fine needle aspiration, highlighting cellular morphology and abnormalities.
(4) IHC (Immunohistochemistry): Stained tissue sections demonstrating protein expression using labeled antibodies, aiding in disease classification and biomarker identification.
(5) Ultrasonography: Diagnostic images acquired via ultrasound, often used for evaluating soft tissues and internal organs.
(6) CT/MRI: Radiological images generated from computed tomography or magnetic resonance imaging, providing detailed cross-sectional anatomical views.
(7) X-ray: Projectional radiographs commonly employed to assess skeletal, thoracic, or abdominal regions.
(8) Schematic diagram: Hand-drawn or computer-generated illustrations used to convey anatomical, physiological, or pathological concepts.
(9) Table: Structured tabular representations summarizing clinical findings, diagnostic criteria, treatment plans, or comparative data.
(10) Other: Images not falling into the above categories, including but not limited to flowcharts, 3D reconstructions, or hybrid multimodal visualizations.
(11) FISH (Fluorescence In Situ Hybridization): Fluorescent microscopy images showing the hybridization of labeled DNA or RNA probes to specific genomic regions, commonly used for detecting chromosomal abnormalities and gene amplifications. This subfield is manually inspected in the manual inspection after the initial classification.

\subsubsection{Tissue System Classification} 
\label{tissuesystemcls}
In the classification of tissue systems in pathology based on H\&E staining, pathological specimens are usually divided into several major categories according to different organs and systems \cite{conch}. These include: (1) Skin and hair: diseases involving the skin and its appendages. (2) Female reproductive system: including related tissues such as the uterus, ovaries, and fallopian tubes. (3) Urinary and male reproductive system: covering organs such as the kidneys, bladder, and prostate. (4) Endocrine system: diseases involving glandular and endocrine functions. (5) Oral and head and neck: including lesions in the oral cavity, throat, and larynx. (6) Bone and soft tissue: including pathological changes in bones, joints, and soft tissues. (7) Infectious diseases: involving infections caused by bacteria, viruses, fungi, etc. (8) Central nervous system: diseases involving the brain, spinal cord, and other parts of the nervous system. (9) Eyes, ears, and nose: including lesions of related organs such as vision, hearing, and smell. (10) Neonatal, Pediatric, and Childhood Diseases: diseases specifically for children of different ages. (11) Transplanted tissue: involving pathological reactions after organ transplantation. (12) Blood, lymph, bone marrow and spleen: mainly includes pathological changes of the blood system and lymphatic system. (13) Gastrointestinal tract and hepatobiliary and pancreatic digestive system: involves diseases of digestive system related organs such as stomach, intestines, liver, gallbladder, pancreas, etc. (14) Breast: includes breast and related diseases. (15) Chest: covers lesions of the lungs, heart and other chest organs.
\subsubsection{Gross System Classification} \label{grosssystemcls}
The classification of pathological specimens can vary slightly when considering macroscopic categories compared to the H\&E staining-based classification \cite{pathgen16m}. The macroscopic classification typically divides specimens into the following major categories: (1) Cardiovascular System: Diseases affecting the heart and blood vessels. (2) Skin and Adnexa: Includes the skin and its appendages, such as hair, nails, and sweat glands. (3) Breast: Pathologies related to breast tissue, including benign and malignant conditions. (4) Head and Neck: Disorders of the head and neck region, excluding the central nervous system, such as the mouth, pharynx, and larynx. (5) Gastrointestinal Tract: Diseases affecting the stomach, intestines, and related structures. (6) Hepatobiliary, Biliary System, and Pancreas: Pathological conditions affecting the liver, bile ducts, and pancreas. (7) Urinary System and the Male Genital Tract: Includes the kidneys, bladder, prostate, and other male reproductive organs. (8) Female Reproductive Tract: Pathologies affecting the uterus, ovaries, fallopian tubes, and other female reproductive organs. (9) Respiratory Tract: Conditions of the lungs, trachea, and bronchi. (10) Bone and Soft Tissue: Disorders of bones, joints, and soft tissues like muscles and fat. (11) Hematopoietic and Lymphatic Tissues: Pathologies related to the blood-forming tissues, including bone marrow and lymph nodes. (12) Endocrine System: Diseases affecting endocrine glands such as the thyroid, and adrenal glands. (13) The Eye, Ear, and Nose: Includes pathologies related to the eyes, ears, and nose. (14) Dental and Orofacial: Disorders of the teeth, gums, and oral cavity. (15) Central and Peripheral Nervous System: Conditions affecting the brain, spinal cord, and peripheral nerves.
\subsection{Image-Text Pair Sources} \label{appendix:image-text pair sources}

\subsubsection{Image-Caption Pair Extraction From Textbooks}

We used a two-step image-caption extraction method.

\begin{figure}[htbp]
    \centering
    \begin{minipage}[t]{0.48\textwidth}
        \centering
        \includegraphics[width=\linewidth]{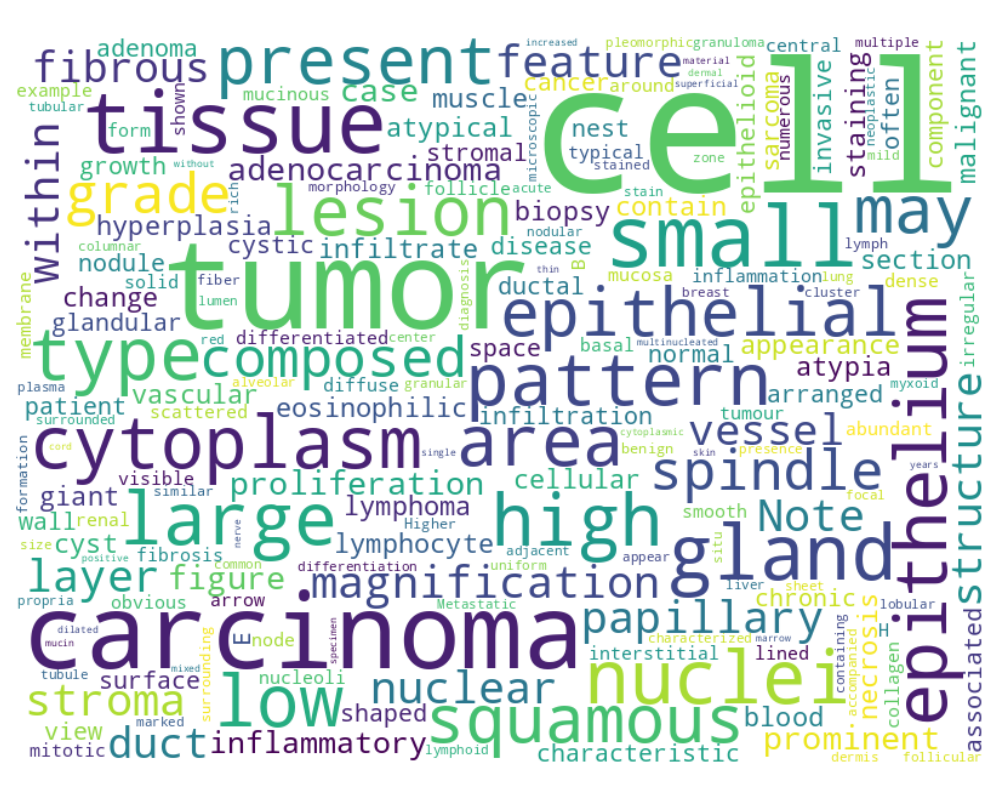} 
        \caption{Word cloud of H\&E} 
        \label{fig:histology} 
    \end{minipage}
    \hfill
    \begin{minipage}[t]{0.48\textwidth}
        \centering
        \includegraphics[width=\linewidth]{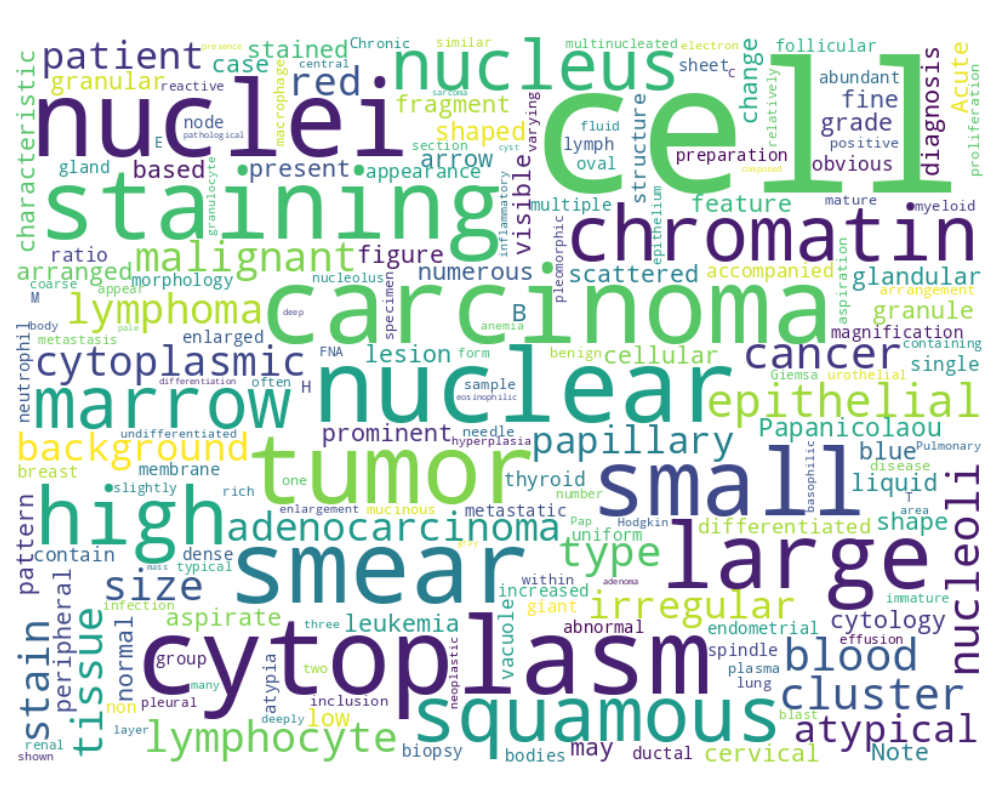} 
        \caption{Word cloud of Cytology}
        \label{fig:cytology}
    \end{minipage}
    
    \vspace{1em} 
    \begin{minipage}[t]{0.48\textwidth}
        \centering
        \includegraphics[width=\linewidth]{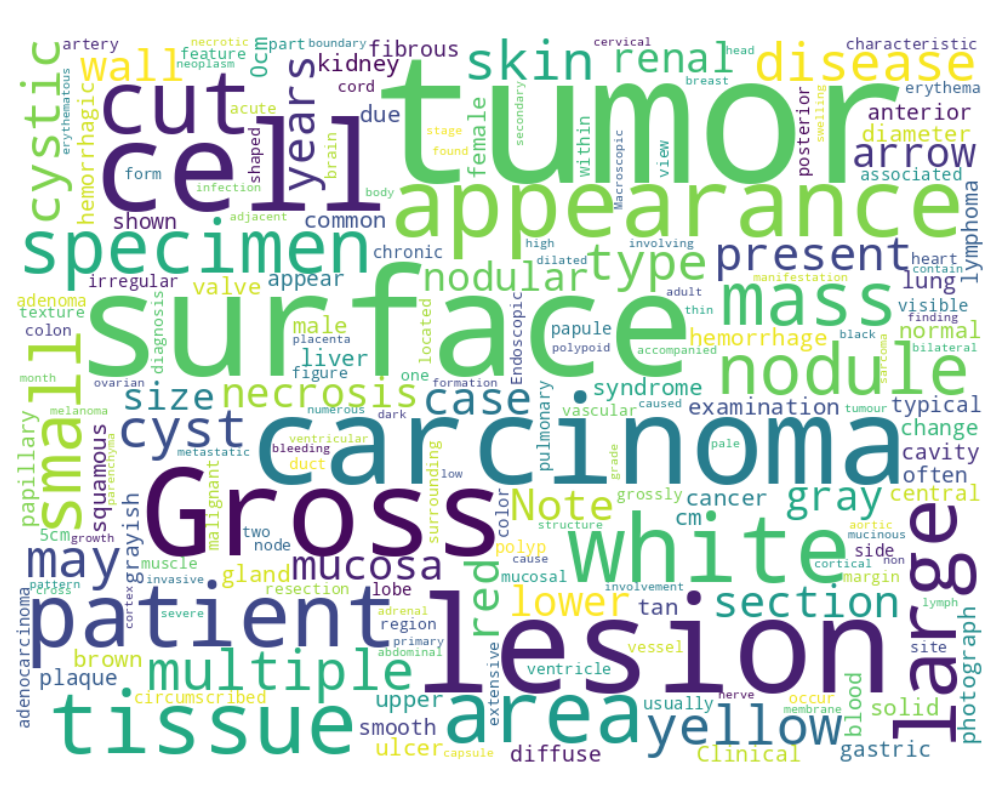} 
        \caption{Word cloud of Gross}
        \label{fig:gross}
    \end{minipage}
    \hfill
    \begin{minipage}[t]{0.48\textwidth}
        \centering
        \includegraphics[width=\linewidth]{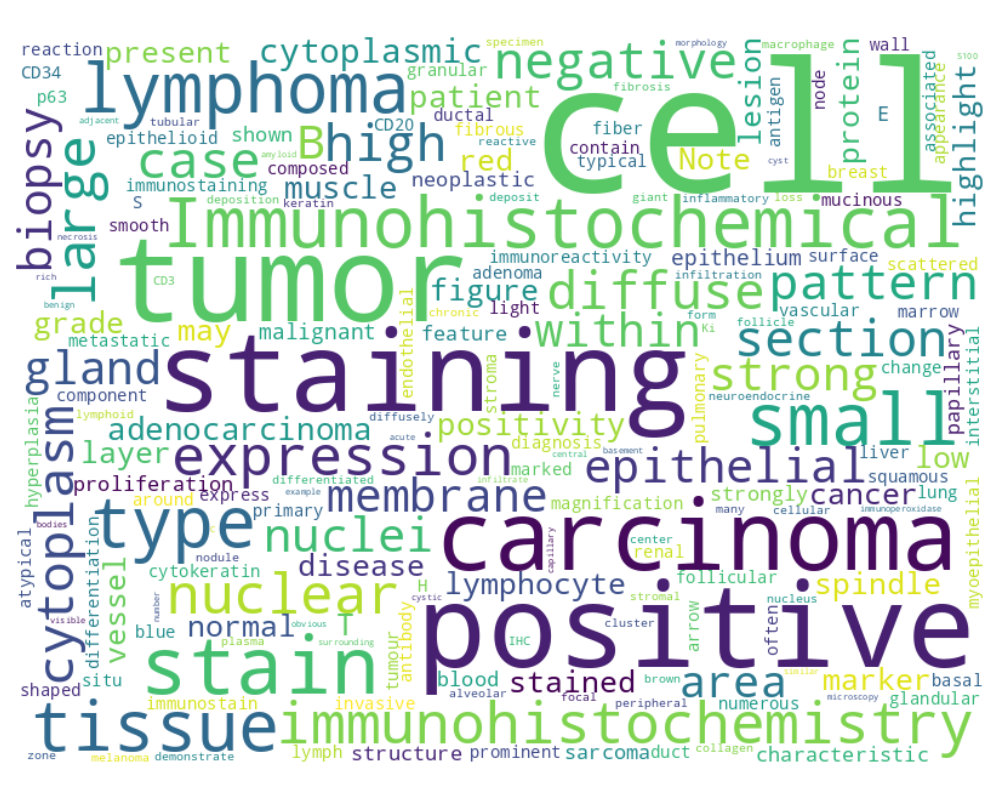} 
        \caption{Word cloud of IHC}
        \label{fig:IHC}
    \end{minipage}
    \label{fig:pathology-images} 
\end{figure}

\textbf{Figure-Caption Pair Extraction.} We performed document layout analysis using \href{https://github.com/opendatalab/DocLayout-YOLO}{DocLayoutYolo}. Each document page was segmented into 7 non-overlapping regions: title, plain text, figure, figure caption, table, table caption, abandon (the running head that contains the folio and the chapter title). We only focus on the plain text, figure, and figure caption region. For figure and figure caption pairing, we performed matching based on spatial proximity, where the spatial distance between any two regions is defined as the vertical distance between their closest bounding box edges. Caption and plain text regions were subsequently recognized using \href{https://github.com/PaddlePaddle/PaddleOCR}{PaddleOCR} to extract textual data. Data extracted from Chinese textbooks is translated into English using Llama-3-7B model after OCR. We also analyze the word frequency distribution within our dataset using word clouds. Specifically, we provide word clouds for four pathological subfields: H\&E~\ref{fig:histology}, Cytology~\ref{fig:cytology}, Gross~\ref{fig:gross}, and IHC~\ref{fig:IHC}.
\textbf{Multi-Panel Figure Cropping.} For multi-panel figures, we detected white boundaries in panels to crop composite images into individual sub-figures. Then, we detected panel labels (e.g., A, B, C) that often appear in corners of figures using text detection, and recognized them with PaddleOCR. We then applied a rule-based composite caption parsing algorithm, which detected pre-defined sub-figure labels of different styles (e.g., A, B, C, or (a), (b), (c)) in alphabetic sequence, and then aligned with each panel. Content appearing before the first panel label within the caption was treated as shared content and was uniformly added to the beginning of each panel caption.

\textbf{In-line References Extraction.} One key advantage of textbook data lies in the rich semantic information found not only in figure captions but within the surrounding narrative of the body text. The clinically relevant content, such as diagnostic indicators, common pathological manifestations, provides deeper insight into the disease mechanisms, enabling models to move beyond surface-level visual-text alignment toward more meaningful pathological understanding. To extract this information, we first converted body text regions into text using PaddleOCR. To account for the possibility that some in-line references may appear on pages different from their associated figures, we consider the textual content within a three-page window centered around each figure. To minimize human effort, we used Qwen-max model to identify in-line references by providing it with each figure caption together with the surrounding text in adjacent pages. 

\textbf{A Practical Alternative.} Besides the method described above, we also experimented with a more convenient but less cost-efficient alternative: Qwen2.5-VL-72B, which has pixel-level spatial understanding and is particularly well-suited for the image-caption extraction task. For books with more complex structures, where our method struggled, we applied Qwen2.5-VL-72B via API calls for document parsing and image-caption extraction. We did not use this method more extensively primarily due to its high computational cost, and also because by the time of its release, most of our figure-caption extraction had already been completed. However, Qwen2.5-VL-72B is highly recommended if budget computational resources are sufficient and high accuracy in document parsing is a priority. 

\subsubsection{Image-Caption Pair Collection From Public Datasets}
We also leverage several large-scale public datasets to construct more image-caption pairs for pathology-focused multimodal learning.

\textbf{The PathCap Dataset} contains approximately 223K image-caption pairs, among which 207K are high-quality pathology-specific examples \cite{pathasst}. The majority of its content is derived from the PubMed Central Open Access Subset, offering a diverse range of medical images spanning multiple pathology subspecialties. Each image is paired with a detailed, domain-specific caption. 

\textbf{The Quilt-1M Dataset} primarily consists of approximately 600K pathology-related images and one million text descriptions, with many images linked to multiple captions \cite{quilt1m}. Unlike traditional academic sources, this dataset draws primarily from social media platforms such as YouTube and Twitter, supplemented by filtered samples from the LAION dataset. As such, Quilt-1M provides a complementary perspective to scholarly datasets by capturing more diverse and real-world usages of medical visuals, though with greater variability in quality. 

\textbf{The PathGen-1.6M Dataset} contains over 1.6 million high-quality image-text pairs, making it currently one of the largest and most refined datasets in the pathology domain \cite{pathgen16m}. The majority of this dataset is sourced from The Cancer Genome Atlas platform, a comprehensive, publicly funded project that provides clinical, genomic, and pathology data across various cancer types.

\subsection{CoT-SFT Data Generation Scheme} \label{appendix:cot-sft data generation scheme}

\subsubsection{Subfield Classification}

We classified the image-text pairs into initially 11 subfields as described in Appendix~\ref{sec:a12}, of which 10 subfields were classified by an EfficientNet-B4 backboned classifier, and `FISH' is identified and manually collected during post-hoc manual inspection. The distribution of samples across the categories is as follows: Histopathology (352,071), Cytology (41,582), Gross Examination (78,177), Immunohistochemistry (IHC) (100,226), CT/MRI (16,713), Ultrasonography (2,594), Schematic Diagram (29,790), Table (3,420), X-ray (13,343), FISH (816), Others (9,390). In addition, we collected 52,136 integrated image-text pairs from expert pathologists’ notes, which were independently compiled during their routine clinical practice.

\subsubsection{Level Clustering}

\begin{figure}[h]
    \centering
    \includegraphics[width=1\linewidth]{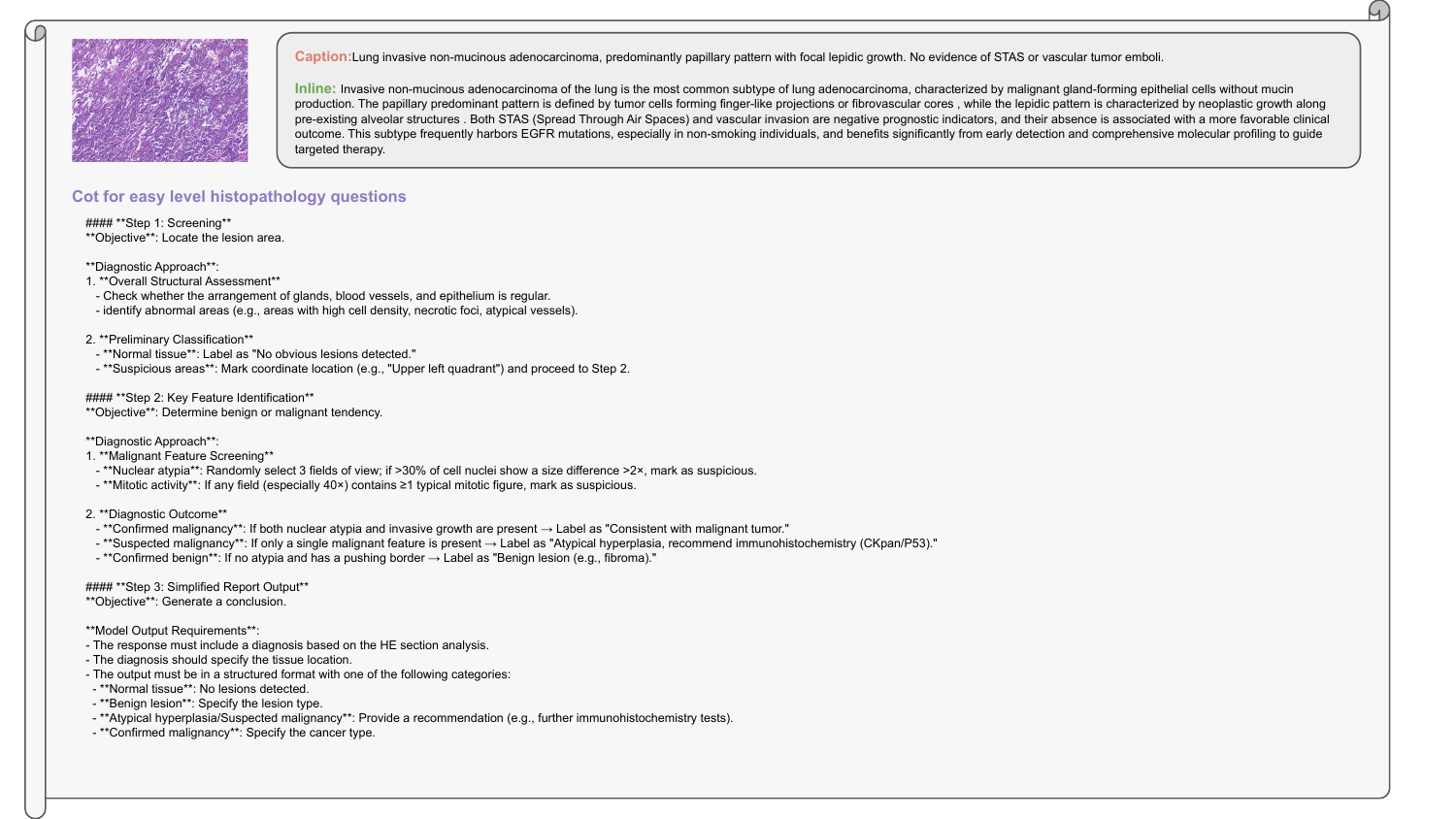}
    \caption{Caption and inline reference for the center of easy-level histopathology questions, followed by the CoT for all easy-level histopathology questions.}
    \label{easy-cot}
\end{figure}

\begin{figure}
    \centering
    \includegraphics[width=1\linewidth]{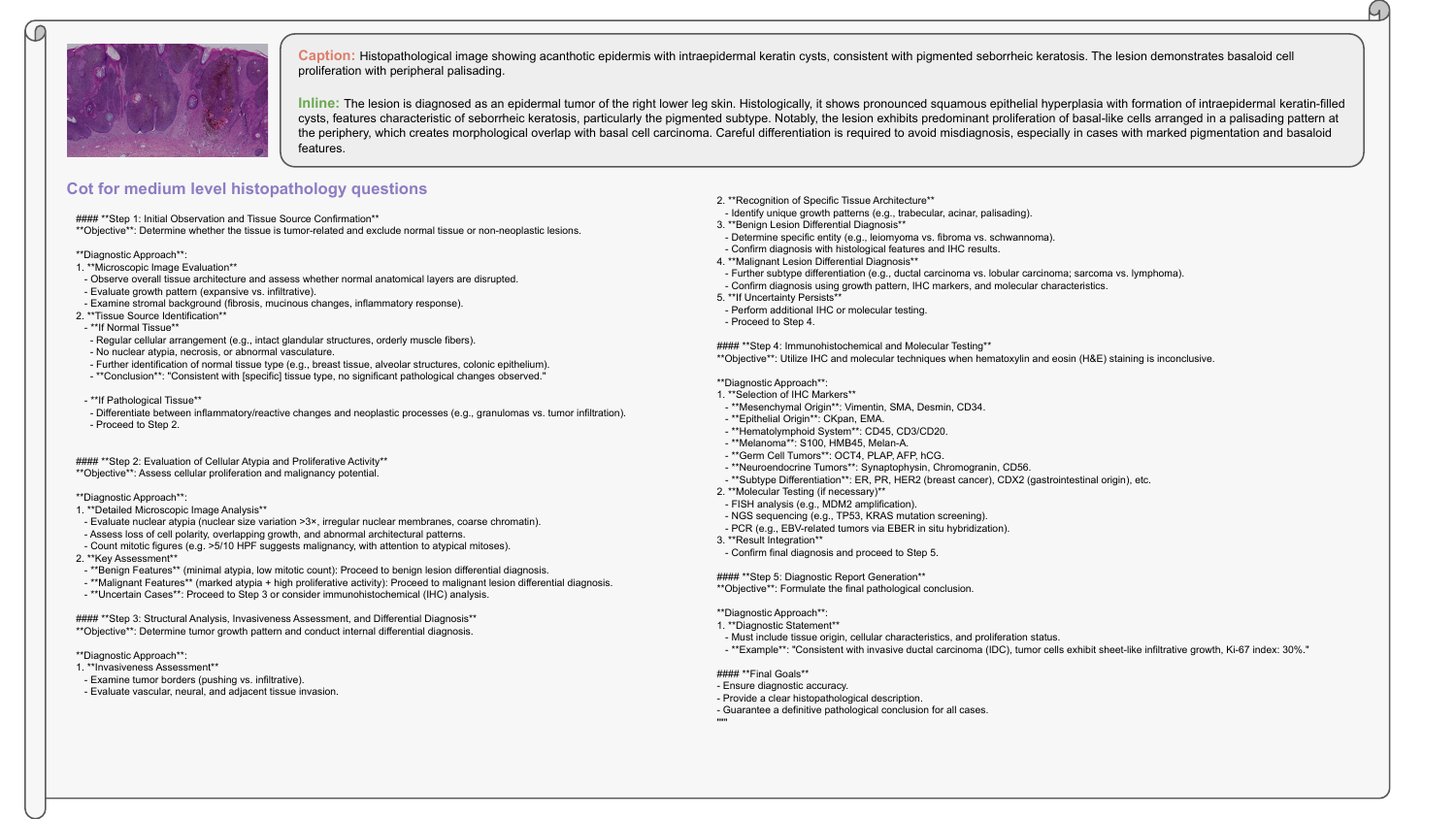}
    \caption{Caption and inline reference for the center of medium-level histopathology questions, followed by the CoT for all medium-level histopathology questions.}
    \label{medium-cot}
\end{figure}

\begin{figure}
    \centering
    \includegraphics[width=1\linewidth]{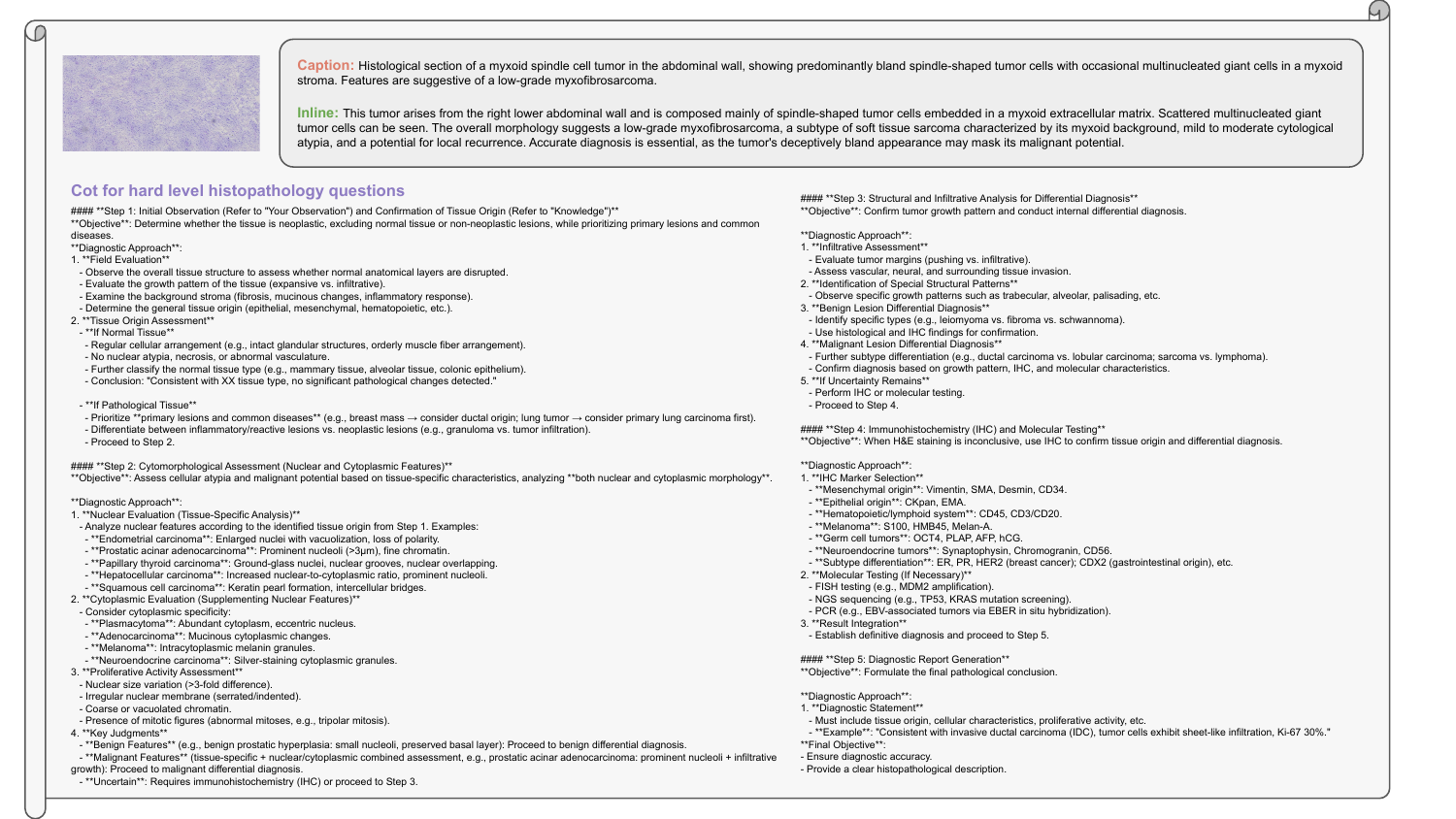}
    \caption{Caption and inline reference for the center of hard-level histopathology questions, followed by the CoT for all hard-level histopathology questions.}
    \label{hard-cot}
\end{figure}

To address the dual requirements of diversity and difficulty balance in SFT data construction, we further categorized a dataset comprising 500K instances spanning five subfields. Our categorization strategy involved generating joint embeddings of both images and accompanying textual content, then applied a three-way clustering algorithm based on these embeddings. Analyzing the resulting cluster centroids revealed significant differences in inherent complexity, enabling us to assign the clusters into three difficulty levels: easy \ref{easy-cot}, medium \ref{medium-cot}, and hard \ref{hard-cot}. This resulted in a total of 15 distinct subcategories across all subfields. As an illustrative example, we present a level-specific cot from histopathology subfield.


\newpage

For downstream data utilization, we adopted a targeted prompting strategy based on the difficulty and reference characteristics of each category. Specifically, samples lacking in-line references were designated for descriptive generation tasks, as these typically require more direct explanations. For the remaining samples, we randomly allocated 60\% to complex reasoning tasks, 30\% to multiple-choice question generation, and 10\% to long-form conversational tasks. 

\begin{tcolorbox}[
  colback=bodybg,        
  coltext=black,         
  colframe=black,        
  title=Prompt used to generate detailed description,
  coltitle=white,        
  colbacktitle=titlebg,  
  fonttitle=\bfseries,   
  fontupper=\ttfamily,
  boxrule=0.3mm,         
  arc=2mm,               
  enhanced
]

Your task is to generate a \textbf{single round of diagnostic reasoning dialogue} based on your observation of the image:  
\\

\#\#\# Instructions:
 \\
 
- You would observe the image first.\\

- You will be provided with a Chain of Thought, which offers some guidance on how to analyze the images. \\

- Do not use phrase like "mentioned", "context", "described", "given" or "provided" in the dialogue. Instead, refer to the information as being "in the image".\\

- Based on your observation of the the image and the Chain of Thought, generate a single round of dialogue.\\

- The dialogue contains a user's question, which you will sample properly from the given possible question list, and a expert's thinking process (analyzed using the Chain of Thought) and a final answer.\\

- The output should be formatted according to the ** response format part.\\

\#\#\# Inputs: \\

- ** Your observation: \\
\{caption\} \\

\#\#\# Chain of thought: \\

\{cot\}  \\

\#\#\# User Question: \\

- The user will ask questions such as: \\
\{question\} \\

\#\#\# response format: \\

User: One of the questions mentioned above. \\

Expert: \\
<think>think step by step based on the chain of thought without skipping any subpoints, summarizing your reasoning into a single, well-structured paragraph. </think> \\
<answer> response followed by the thinking </answer>
 \\

\#\#\# Example:\\

\{example\}

\end{tcolorbox}

\begin{tcolorbox}[
  colback=bodybg,        
  coltext=black,         
  colframe=black,        
  title=Prompt used to generate complex reasoning,
  coltitle=white,        
  colbacktitle=titlebg,  
  fonttitle=\bfseries, 
  fontupper=\ttfamily,
  boxrule=0.3mm,         
  arc=2mm,               
  enhanced
]

Your task is to generate a \textbf{single round of diagnostic reasoning dialogue containing a user's question and a expert's thinking process and answer}. Complete the thinking process based on your observation of the image and related knowledge, then provide a clear answer. \\

\#\#\# Instructions: \\

- You would observe the image first and recall some related knowledge.\\

- You will be provided with a Chain of Thought, which offers some guidance on how to analyze the images.\\

- Do not use phrase like "mentioned", "context", "described", "given" or "provided" in the dialogue. Instead, refer to the information as being "in the image".\\

- Based on your observation of the image, related knowledge and the Chain of Thought, generate a single round of dialogue.\\

- The dialogue contains a user's question, which you will sample properly from the given possible question list, and a expert's thinking process (analyzed using the Chain of Thought) and a final answer.\\

- The output should be formatted according to the ** response format part. \\

\#\#\# Inputs: \\

- \textbf{Your observation:} \\
\{caption\} \\
- \textbf{Knowledge:} \\
\{inline\} \\

\#\#\# Chain of thought: \\

\{cot\} \\

\#\#\# User Question: \\

- The user will ask questions such as: \\
\{question\} \\

\#\#\# response format: \\

User: One of the questions mentioned above. \\

Expert: \\

<think>think step by step based on the chain of thought without skipping any subpoints, summarizing your reasoning into a single, well-structured paragraph.</think> \\
<answer> response followed by the thinking </answer> \\

{\large \textbf{Example:}} \\

\{example\}

\end{tcolorbox}

\begin{tcolorbox}[
  colback=bodybg,        
  coltext=black,         
  colframe=black,        
  title=Prompt used to generate MCQs,
  coltitle=white,        
  colbacktitle=titlebg,  
  fonttitle=\bfseries,   
  fontupper=\ttfamily,
  boxrule=0.3mm,         
  arc=2mm,               
  enhanced
]

Your task is to generate a \textbf{multiple-choice question based on your observation and related knowledge of the image, and provide an answer containing the thinking process}. \\

\#\#\# Instructions: \\

- You would observe the image first and recall some related knowledge.\\

- You will be provided with a Chain of Thought, which offers some guidance on how to analyze images.\\

- Do not use phrase like "mentioned", "context", "described", "given" or "provided" in the dialogue. Instead, refer to the information as being "in the image".\\

- Based on your observation, related knowledge and the Chain of Thought, generate a multiple-choice question.\\

- \textbf{Ensure the question is clinically meaningful} for pathology learners.\\

- Your question should contain 4 choices, including 3 wrong answers and 1 correct answer.\\

- The output should be formatted according to the \textbf{output format} part.\\

\#\#\# Inputs: \\

- \textbf{Your observation:} \\
\{caption\} \\
- \textbf{Knowledge:} \\
\{inline\} \\

\#\#\# Chain of thought: \\

\{cot\} \\

{\textbf{Output Format:}} \\

Question: Your question here \\
A) Option A \\
B) Option B \\
C) Option C \\
D) Option D \\

<think>think step by step based on the chain of thought without skipping any subpoints, and summarize your reasoning into a single, well-structured paragraph.</think> \\
<answer> Letter of the Correct Answer </answer> \\

{\large \textbf{Example:}} \\

\{example\}
\end{tcolorbox}

\begin{tcolorbox}[
  colback=bodybg,
  coltext=black,
  colframe=black,
  title=Prompt used to generate 3-round diagnostic dialogue,
  coltitle=white,
  colbacktitle=titlebg,
  fonttitle=\bfseries,
  fontupper=\ttfamily,
  boxrule=0.3mm,
  arc=2mm,
  enhanced
]

Your task is to generate a \textbf{3-round dialogue} between a user and an expert. The conversation should progress in depth, where each user question builds on the previous one, leading to a more detailed or complex inquiry. Only the last (third) round of expert response should include a full thinking process using the Chain of Thought and answers.

\#\#\# Instructions: \\

- You would observe the image first and recall some related knowledge. \\

- Generate a 3-turn conversation (user-expert-user-expert-user-expert). Each user question should be more specific and detailed than the previous one.\\

- Do not use phrase like "mentioned", "context", "described", "given" or "provided" in the dialogue. Instead, refer to the information as being "in the image".\\

- The first two rounds should be simple Q\&A based on the image and knowledge, without a formal thinking process.\\

- In the last round, expert must reason through the problem using the provided Chain of Thought and provide a clear answer.\\

\#\#\# Inputs: \\

- \textbf{Your observation:} \\
\{caption\} \\
- \textbf{Knowledge:} \\
\{inline\} \\

\#\#\# Chain of thought: \\

\{cot\} \\

{\textbf{Output Format:}} \\

<Round 1> \\
User: \textless simple question 1\textgreater \\
Expert: \textless brief factual answer\textgreater \\
<Round 2> \\
User: \textless question 2 based on previous answer, slightly deeper\textgreater \\
Expert: \textless brief factual answer\textgreater \\
<Round 3> \\
User: \textless question 3 that requires reasoning\textgreater \\
Expert: \\
<think> \\
Step-by-step reasoning based on the chain of thought \\
</think> \\
<answer> \\
A clear conclusion derived from the thinking \\
</answer> \\

{\large \textbf{Example:}} \\

\{example\}
\end{tcolorbox}

\subsection{RL Dataset Construction} \label{appendix:rl dataset construction}

In the process of constructing the reinforcement learning dataset, we created a multiple-choice question dataset comprising 10,000 questions, each with six answer options. The composition was as follows: 5,000 H\&E samples, 1,000 Gross pathology samples, 1,900 Cytology samples, 1,900 IHC samples, and 200 FISH samples. Unlike conventional MCQ formats that typically use four options, the expanded six-option format was intentionally adopted to increase task difficulty and reduce the likelihood of random guessing, thereby better aligning with the requirements of RL training, which benefits from more discriminative reward signals. This dataset is entirely independent of the MCQ dataset used in the supervised fine-tuning phase.

To ensure diagnostic focus and eliminate potential biases from retrieval-based reasoning, we excluded in-line reference information in the RL MCQ dataset. For each subfield, we selected samples with the longest caption to maximize the richness of clinical information.

\begin{figure}[htb]
    \centering
    \includegraphics[width=1\linewidth]{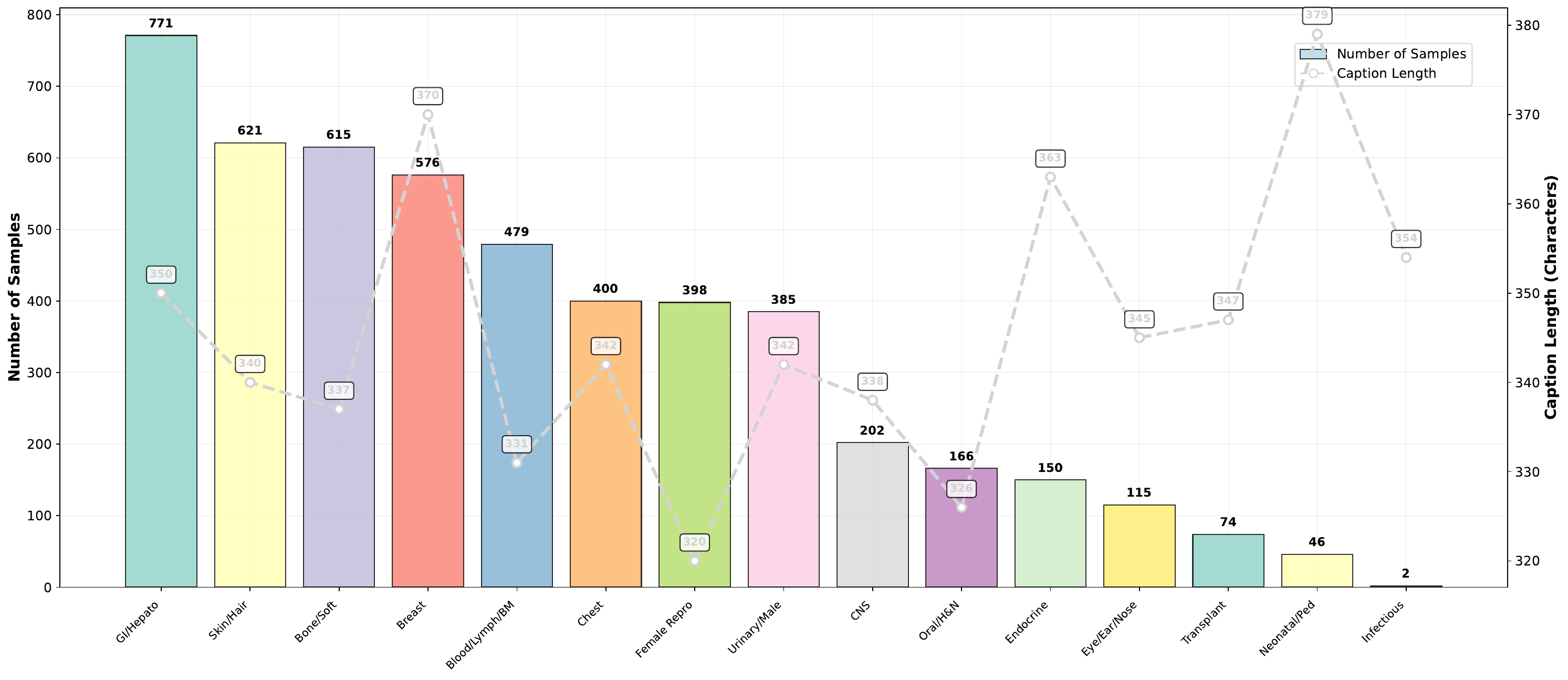}
    \caption{Based on the tissue system classification, the composition distribution of the MCQs dataset of H\&E and the minimum caption length of each category}
    \label{fig:rldataset_he}
\end{figure}

For the H\&E dataset, we first divided the data into 15 categories based on tissue systems, as defined in Appendix~\ref{tissuesystemcls}. From this categorization, 6,051 samples were proportionally selected according to the tissue system distribution in the original corpus. These were subsequently reviewed by pathologists, who retained the top 5,000 highest-quality samples, ensuring both diversity and diagnostic relevance (see Figure~\ref{fig:rldataset_he}).

In the Gross subset, samples were similarly categorized into 15 gross systems following Appendix~\ref{grosssystemcls}. The total number of samples was then scaled to 1,000 while preserving the original class distribution. The final allocation is as follows: Bone and Soft Tissue (124), Gastrointestinal Tract (128), Cardiovascular System (97), Skin and Adnexa (95), Urinary System and Male Genital Tract (94), Female Reproductive Tract (91), Central and Peripheral Nervous System (73), Respiratory Tract (73), Hepatobiliary System, Biliary System, and Pancreas (56), Head and Neck (39), Endocrine System (38), Eye, Ear, and Nose (27), Hematopoietic and Lymphatic Tissues (26), Breast (22), and Dental and Orofacial (17).

For the remaining three subfields—Cytology, IHC, and FISH—we followed the same sample selection principles (e.g., caption length maximization and quality control), but without further subclassification.

\section{Details of Models and Training}

This section outlines the training procedures for both PathoCLIP and Patho-R1. PathoCLIP uses OpenAI-CLIP as initialization, trained in two stages with different datasets. Patho-R1's training includes CPT and SFT using Qwen2.5-VL, followed by reinforcement learning with GRPO and DAPO algorithms. 

\subsection{CLIP Training} \label{appendix:clip training}
For the PathoCLIP training, we adhere to the \href{https://github.com/mlfoundations/open_clip}{open-clip} framework and use OpenAI CLIP (OpenAI-CLIP-B for PathoCLIP-B, OpenAI-CLIP-L-336 for PathoCLIP-L) as initialization. We use a learning rate of 1e-4 with Adamw optimizer that includes weight decay of 0.05. We set a batch size of 128 across 8 NVIIDA A800 GPUs, resulting in an effective batch size of 1024. As mentioned in the previous chapter ,in the first stage of training using PathGen-1.6M, we limit the training to only one epoch. For the second stage of training with other datasets, including Quit-1M, PathCap and our private dataset, we conduct 10 epochs.

\subsection{LVLM Training} \label{appendix:lvlm training}

\subsubsection{CPT and SFT}
All training below was conducted on 8 NVIDIA A800 GPUs with a per-device batch size of 32, resulting in an effective batch size of 256. (For DAPO, we set dynamic generation batch 256, but training batch size 128.)

\textbf{For Continued Pretraining.} We followed the \href{https://github.com/modelscope/ms-swift.git}{ms-swift} framework and initialized the model with Qwen2.5VL. We used the AdamW optimizer with a learning rate of 5e-5, $\beta_1 = 0.9$, $\beta_2 = 0.95$, and a weight decay of 0.1. The model was trained for one epoch on 3.5 million samples.

\textbf{For Supervised Finetuning.} We adopted the \href{https://github.com/hiyouga/LLaMA-Factory.git}{LLaMA-Factory} framework and froze the vision tower. We used a learning rate of 1e-5 and trained on 500K samples for 3 epochs.

\subsubsection{Reinforcement Learning}

\textbf{Implementation Details.} We adopted the \href{https://github.com/volcengine/verl}{verl} framework for reinforcement learning. For both GRPO and DAPO, we set the actor and critic learning rates 1e-6 and 1e-5 respectively. For GRPO, we trained on 10k samples for 5 epochs. For DAPO, we trained on the same 10k samples for 3 epochs.

\textbf{GRPO and DAPO Algorithm.} For each training iteration, GRPO samples $\mathrm{G}$ candidate output $\{o_i\}_{i=1}^G$ from the old policy $\pi_{\mathrm{old}}$. Each output receives a reward $r_{i}$, from which we compute the group-relative advantage $A_{i}$:
\begin{equation}A_{i}=\frac{r_{i}-\mu_{\{r\}}}{\sigma_{\{r\}}},\end{equation}
where $\mu_{\{r\}}$ and $\sigma_{\{r\}}$ denote the mean and standard deviation fo reward values within the group. Here, $A_{i}$ denotes the group-relative advantage, which emphasizes outputs that surpass the group average for prioritized optimization. The policy is optimized by maximizing the following objective function: 
\begin{equation}\begin{aligned}\mathcal{J}_{\mathrm{GRPO}}(\theta)&=\mathbb{E}_{\mathbf{v}\sim P(\mathbf{V}),\{o_{\mathbf{i}}\}_{\mathbf{i}=1}^{G}\sim\pi_{\theta_{\mathbf{old}}}(O|\mathbf{v})}\\&\left[\frac{1}{G}\sum_{i=1}^{G}\frac{1}{|o_i|}\sum_{t=1}^{|o_i|}\left(\min\left(r_{i,t}A_{i,t},\mathrm{clip}\left(r_{i,t},1-\epsilon,1+\epsilon\right)A_{i,t}\right)-\beta D_{\mathrm{KL}}(\pi_{\theta}\parallel\pi_{\mathrm{ref}})\right)\right],
\end{aligned}
\end{equation}

where $\mathbf{v}$ represents a conditioning input drawn from a distribution $P(\mathbf{V})$; $r_{i,t} = \frac{\pi_{\theta}(o_{i,t} \mid \mathbf{v})}{\pi_{\theta_{\text{old}}}(o_{i,t} \mid \mathbf{v})}$ represents the importance sampling ratio that reflects the relative likelihood of generating output $o_{i,t}$ under the new policy $\pi_\theta$ compared to the old policy $\pi_{\theta_{\mathbf{old}}}$; the clipping operator clip $(r_{i,t},1-\epsilon,1+\epsilon)$ constrains updates to remain within a trust region $[1-\epsilon, 1+\epsilon]$ to prevent excessively large policy shifts, with $\epsilon$ being a hyperparameter. By taking the minimum between the unclipped and clipped terms, the objective encourages stable yet effective updates. The final term, $D_{\mathrm{KL}}(\pi_{\theta}\parallel\pi_{\mathrm{ref}})$, regularizes the policy by penalizing divergence from a reference policy, weighted by a coefficient $\beta$.
In addition, we adopted the DAPO algorithm to further investigate the comparative effectiveness of different policy optimization strategies.
DAPO enhances the policy optimization process by sampling a group of outputs $\{o_i\}_{i=1}^G$ for each conditioning input and optimizing the policy through a modified token-level objective function:
\begin{equation}
\begin{aligned}
\mathcal{J}_{\mathrm{DAPO}}(\theta) &= \mathbb{E}_{\mathbf{v}\sim P(\mathbf{V}),\{o_{\mathbf{i}}\}_{\mathbf{i}=1}^{G}\sim\pi_{\theta_{\mathbf{old}}}(O|\mathbf{v})} \\
&\left[\frac{1}{\sum_{i=1}^{G}|o_i|}\sum_{i=1}^{G}\sum_{t=1}^{|o_i|}\min\left(r_{i,t}A_{i,t},\mathrm{clip}\left(r_{i,t},1-\epsilon_{\mathrm{low}},1+\epsilon_{\mathrm{high}}\right)A_{i,t}\right)\right], \\
\end{aligned}
\end{equation}

subject to the constraint:

\begin{equation}
0 < |\{o_i | \mathrm{is\_equivalent}(answer, o_i)\}| < G,
\end{equation}

where $\epsilon_{\mathrm{low}}$ and $\epsilon_{\mathrm{high}}$ define asymmetric clipping bounds to stabilize training by limiting policy deviation, $r_{i,t}$ denotes the token-level importance sampling ratio and $A_{i,t}$ is the group-normalized advantage, computed by standardizing the return $r$ across the group.

\textbf{Reward Function Design}
We design reward functions to guide model learning from structured and accurate responses. For GRPO training, we employ two types of reward signals: format reward and accuracy reward. For DAPO, we further introduce a length-aware penalty to discourage overly long responses.

\textbf{Format Reward}
To encourage structured reasoning and output readability, we require the model to format its response using <think>...</think> for intermediate reasoning steps and <answer>...</answer> for the final answer. A reward score of 1 is assigned if both tag pairs appear exactly once and no content exists outside these tags; otherwise, the score is 0. The reward function is defined as:

\begin{equation}
R_{\mathrm{fmt}}(y) =
\begin{cases}
1, & \text{if the format is strictly matched} \\
0, & \text{otherwise}
\end{cases}
\end{equation}

where $y$ denotes the model's output. This format structure helps the model organize its reasoning clearly and improves interpretability.

\textbf{Accuracy Reward}
To evaluate the correctness of the final answer, we adopt a strict multiple-choice matching strategy. The content within the <answer>...</answer> tags is extracted using regular expressions, and only exact matches to a valid option from the predefined set {A, B, C, D, E, F} are accepted. A reward score of 1 is assigned if the extracted answer matches the ground-truth label; otherwise, the score is 0. The reward function is defined as:

\begin{equation}
R_{\mathrm{acc}}(y, y^*) =
\begin{cases}
1, & \text{if } y \in {A, B, C, D, E, F} \text{ and } y = y^* \\
0, & \text{otherwise}
\end{cases}
\end{equation}

where $y$ is the model's predicted answer extracted from the <answer>...</answer> tag, and $y^*$ is the reference correct choice. This design ensures accurate, format-consistent outputs suitable for automated scoring and reinforces clarity in response generation.

\textbf{Length-Aware Penalty}
During the DAPO training process, we incorporate a Soft Overlong Punishment to discourage excessively long outputs. If a response exceeds the predefined maximum length $L_{\text{max}}$, we apply a continuous penalty within a soft margin interval of length $L_{\text{cache}}$, and a hard penalty beyond $L_{\text{max}}$. The length penalty is defined as:

\begin{equation}
R_{\mathrm{length}}(y)=\begin{cases}
0, & |y|\leq L_{\mathrm{max}}-L_{\mathrm{cache}} \\
\frac{(L_{\mathrm{max}}-L_{\mathrm{cache}})-|y|}{L_{\mathrm{cache}}}, & L_{\mathrm{max}}-L_{\mathrm{cache}}<|y|\leq L_{\mathrm{max}} \\
-1, & L_{\mathrm{max}}<|y|
\end{cases}
\end{equation}

\section{More Experiment Results}

In this section, we present details of the CLIP benchmark, which encompasses zero-shot classification tasks, cross-modal tasks and multimodal benchmarks for LVLMs, including MCQ, YORN and VQA. Additionally, we provide further details on the linear probing experiments.

\subsection{CLIP Evaluation} \label{appendix:clip evaluation}
\subsubsection{CLIP Benchmark}

\textbf{Zero-Shot Classification.} We employ 5 datasets  for tasks related to zero-shot classification, which incude: \textbf{LC-25000} \cite{lung}: This dataset includes samples of lung and colon adenocarcinomas, organized into two subsets: LC-lung, which encompasses lung adenocarcinomas, lung squamous cell carcinomas, and benign lung tissues; and LC-colon, which contains colon adenocarcinomas and benign colonic tissues. \textbf{WSSSLUAD} \cite{wsss}:Featuring patch-level annotations from 87 whole slide images, this collection is focused on distinguishing between tumor and normal tissue classes. \textbf{SICAPv2} \cite{sicapv2}:This dataset features images of prostate pathology magnified 10 times, classified as non-cancerous, and Grades 3-5 according to the Gleason grading system. \textbf{BMT} \cite{bmt}:This dataset is a multicellular ThinPrep® dataset, consisting of 600 clinically reviewed images collected from 180 Pap smears from 180 patients, which are categorized into three key diagnostic classes: normal squamous epithelium, low-grade squamous intraepithelial lesion, and high-grade squamous intraepithelial lesion.Instead of a single label, we designed a collection of multiple text prompts for each category \cite{conch}. For example, for “LUSC” in the lung tumor classification of LC-25000Lung, different expressions such as "squamous cell carcinoma", "lung squamous cell carcinomas","squamous cell carcinoma of the lung", “Pulmonary squamous cell carcinoma” and "SCC of the lung" were used. The reason for the enriched labeling is that there are significant differences in the textual descriptions of the same pathological entity (e.g., full name, abbreviation, etc.) in the pathology reports and in the literature. A single label tends to cause the model to miss semantically equivalent representations during zero-sample matching, reducing the generalization ability. In addition to this, with the prompt ensemble, the model can more comprehensively capture the alignment between images and text in the embedding space of contrastive learning. Table \ref{zeroclass} presents the set of class labels used for each dataset in zero-shot image classification, while Table \ref{suptable:templates} lists the corresponding prompt templates, where CLASSNAME is substituted with the specific class of interest.

\textbf{Cross Modal Retrieval.} We employ 2 datasets for tasks related to cross modal retrieval, which include: \textbf{ARCH} \cite{arch}:The dataset is constructed from PubMed medical articles and pathology textbooks, containing multi-instance images with detailed descriptions across diverse staining methods, tissue types, and pathological diagnoses. For our cross-modal retrieval task, we focus on the book\_set subset of the ARCH dataset. Notably, a minor portion of this subset contains cases where one textual description corresponds to multiple images. To isolate and evaluate the PathoCLIP model's ability to retrieve unique image-text correspondences without ambiguity from multi-image contexts, we filtered out these one-to-many cases. This process yielded a final dataset of 2,720 clean one-to-one image-text pairs, ensuring each description uniquely maps to a single image for precise retrieval performance assessment. \textbf{Archive}:This dataset is a self-constructed picture-description dataset independently developed by us, covering a diverse spectrum of pathological categories. Predominantly sourced from various specialized pathology textbooks, it systematically encompasses nearly all major pathological categories. Specifically designed to evaluate model generalization in cross-modal retrieval tasks, the dataset comprises 3,060 manually annotated image-text pairs, with each pair carefully curated to ensure semantic consistency and domain representativeness.

\begin{table}[ht]
  \centering
  \caption{Classname for each dataset on zero-shot image classification.} 
  \label{tab:zero-shot-classes}
  \begin{tabularx}{\textwidth}{@{} l >{\raggedright\arraybackslash}X @{}}
    \hline
    \bfseries Dataset & \bfseries Classname \\
    \hline

    BMT &
    \textbf{"NIL"}: "Negative", "No atypia", "Benign cervical tissue", "Normal cervical epithelium", "Non-neoplastic cervical tissue", "Uninvolved cervical mucosa", "No HPV-related alterations", "No intraepithelial neoplasia", "No squamous metaplasia" \newline
    \textbf{"LSIL"}: "Low-grade squamous intraepithelial lesion", "Low-grade dysplasia", "Low-grade lesion", "Mild atypical hyperplasia", "Koilocytotic atypia", "Low-grade squamous metaplasia" \newline
    \textbf{"HSIL"}: "High-grade squamous intraepithelial lesion", "Severe cervical dysplasia", "High-grade dysplasia" \\[2.5ex]

    \hline
    LC-25000Lung &
    \textbf{"LUAD"}: "Lung adenocarcinoma", "Adenocarcinoma", "Adenocarcinoma of the lung" \newline
    \textbf{"LUSC"}: "Squamous cell carcinoma", "Lung squamous cell carcinomas", "Squamous cell carcinoma of the lung", "Pulmonary squamous cell carcinoma", "SCC of the lung" \newline
    \textbf{"BLT"}: "Benign lung tissue", "Normal lung parenchyma", "Benign pulmonary tissue", "Non-neoplastic lung tissue" \\[2.5ex]

    \hline
    LC-25000Colon &
    \textbf{"COAD"}: "Adenocarcinoma", "Colon adenocarcinoma", "Colonic adenocarcinoma", "Adenocarcinoma of the colon", "Colorectal adenocarcinoma", "Colonic carcinoma", "Epithelial malignancy of colon", "Colon malignancy" \newline
    \textbf{"BCOT"}: "Benign colonic tissue", "Normal colonic mucosa", "Uninvolved colonic tissue", "Non-neoplastic colonic tissue", "Healthy colon tissue", "Benign tissue of the colon" \\[2.5ex]

    \hline
    SICAPv2 &
    \textbf{"G3"}: "Atrophic well differentiated and dense glandular regions", "Well-formed glands", "Well-differentiated glandular architecture", "Small acini with open lumina", "Back-to-back separated glands", "Tubular glands" \newline
    \textbf{"G4"}: "Cribriform, ill-formed, large-fused and papillary glandular patterns", "Poorly formed glands", "Fused glands", "Cribriform glands", "Glomeruloid structures", "Complex fused glands", "Large irregular glands", "Multipseudoluminal cribriform structures" \newline
    \textbf{"G5"}: "Isolated cells or file of cells, nests of cells without lumina formation and pseudo-rosetting patterns", "Single-cell infiltration", "Solid sheets of tumor cells", "Cords of cells", "Comedo-type necrosis", "Central tumor necrosis", "Anaplastic cell clusters", "Micrometastatic-like nests" \newline
    \textbf{"NC"}: "Benign prostate tissue", "Normal prostate gland", "Healthy prostate tissue", "Uninvolved prostate", "No cancer", "Non-cancerous" \\[2.5ex]

    \hline
    WSSSLUAD &
    \textbf{"Tumor"}: "Lung tumor", "Lung tumor tissue", "Cancerous lesion", "Malignant tumor region", "Tumorous area", "Lung adenocarcinoma tumor", "Adenocarcinoma region", "Lung cancer tissue" \newline
    \textbf{"Normal"}: "Normal", "Normal lung tissue", "Healthy lung parenchyma", "Non-cancerous tissue", "Benign tissue", "Normal alveolar structure", "Intact lung epithelium", "Normal pulmonary tissue" \\[2.5ex]

    \hline
  \end{tabularx}
  \label{zeroclass}
\end{table}

\begin{table}[h]
  \centering
  \caption{Prompt templates for zero‐shot image classification (replace \texttt{CLASSNAME} with the desired class).}
  \label{tab:prompt-templates}
  \begin{tabularx}{\textwidth}{@{} r X @{}}
    \toprule
    \bfseries No. & \bfseries Templates \\
    \midrule
     1 & \{CLASSNAME\}. \\
     2 & a photomicrograph showing \{CLASSNAME\}. \\
     3 & a photomicrograph of \{CLASSNAME\}. \\
     4 & an image of \{CLASSNAME\}. \\
     5 & an image showing \{CLASSNAME\}. \\
     6 & an example of \{CLASSNAME\}. \\
     7 & \{CLASSNAME\} is shown. \\
     8 & this is \{CLASSNAME\}. \\
     9 & there is \{CLASSNAME\}. \\
    10 & a histopathological image showing \{CLASSNAME\}. \\
    11 & a histopathological image of \{CLASSNAME\}. \\
    12 & a histopathological photograph of \{CLASSNAME\}. \\
    13 & a histopathological photograph showing \{CLASSNAME\}. \\
    14 & shows \{CLASSNAME\}. \\
    15 & presence of \{CLASSNAME\}. \\
    16 & \{CLASSNAME\} is present. \\
    17 & an H\&E stained image of \{CLASSNAME\}. \\
    18 & an H\&E stained image showing \{CLASSNAME\}. \\
    19 & an H\&E image showing \{CLASSNAME\}. \\
    20 & an H\&E image of \{CLASSNAME\}. \\
    21 & \{CLASSNAME\}, H\&E stain. \\
    22 & \{CLASSNAME\}, H\&E. \\
    \bottomrule
  \end{tabularx}
  \label{suptable:templates}
\end{table}

\subsubsection{Details of Linear Probing}

\begin{figure}[ht]
    \centering
    \begin{subfigure}[b]{0.48\textwidth}
        \centering
        \includegraphics[width=\textwidth]{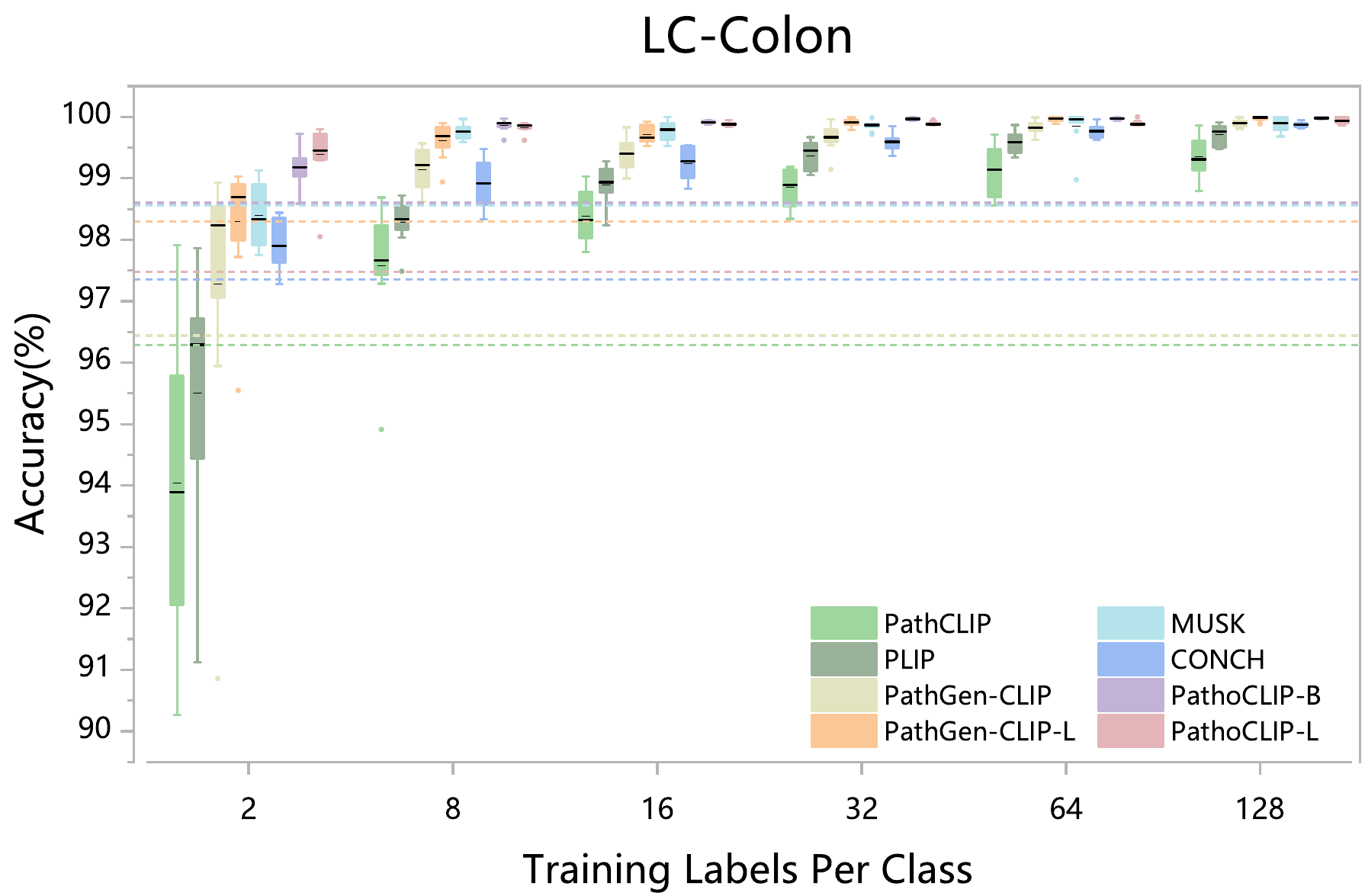}
        \label{fig:lc-lung}
    \end{subfigure}
    \hfill
    \begin{subfigure}[b]{0.48\textwidth}
        \centering
        \includegraphics[width=\textwidth]{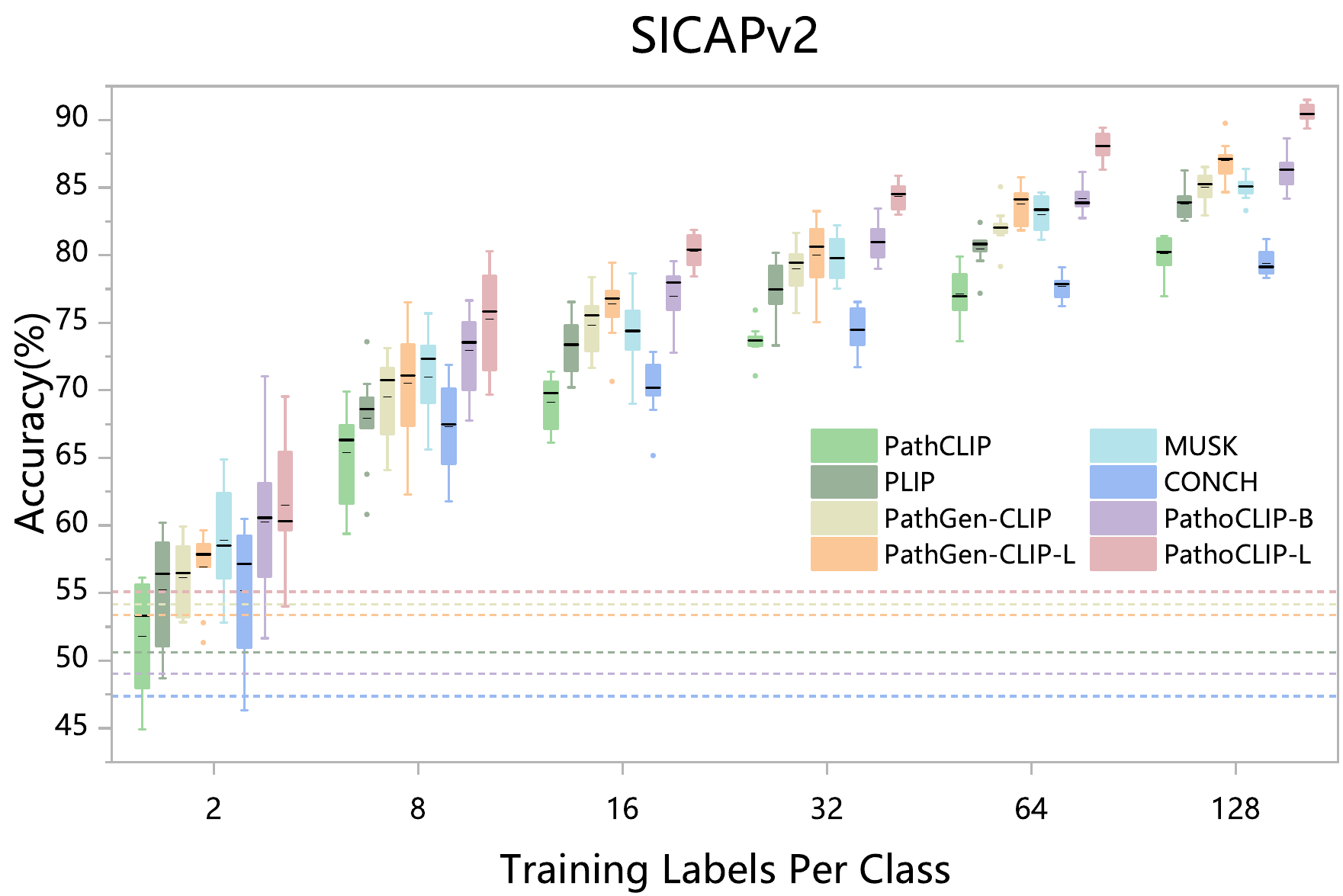}
        \label{fig:bmt}
    \end{subfigure}

    \caption{Few-shot experiments on LC-Colon dataset and SICAPv2 dataset. }
    \label{fig:few_shot_pathology1}
\end{figure}

To evaluate the representational capability of pre-trained models, we introduce a linear probe on their feature representations. Specifically, we freeze the visual encoder of the CLIP model and only train the linear classifier head, aiming to test the model's generalization ability across different disease types and tissue morphologies while maintaining consistency in feature extraction. To comprehensively assess the model performance under low-resource scenarios, we set multiple conditions with annotation counts ranging from 2 to 128 samples for each dataset. For each condition, we train on 10 random subsets of samples to measure the stability and robustness of model performance. During the classifier training phase, we employ multinomial logistic regression as the classifier, setting hyperparameters as C=1.0, solver='lbfgs', and mult\_class='multinomial' to achieve accurate multi-class predictions. In addition to the LC-Lung and BMT datasets described in the main text, we also test on the LC-Colon and SICAPv2 datasets. As shown in Figure \ref{fig:few_shot_pathology1}, the results demonstrate that the proposed method exhibits excellent representativeness and strong semantic generalization ability across various medical image scenarios, which indirectly validates that the features obtained by the model during the pre-training stage possess rich semantic expressiveness.

\newpage
\newpage
\subsection{Multimodal Benchmark for LVLMs} \label{appendix:multimodal benchmark for lvlms}

\subsubsection{MCQ And YORN Benchmark}

For the MCQ dataset, we selected the PathMMU dataset, which is a benchmark specifically designed for pathology. In accordance with the dataset's requirements, we downloaded images sourced from the Twitter platform. It is worth noting that some original posts had been deleted, making the corresponding images unavailable, and thus, those questions had to be excluded. For the PathMMU validation set (PathMMU-val), a total of 705 questions were retained. The distribution across the five subsets is as follows: Atlas – 80, EduContent – 146, PathCLS – 96, PubMed – 233, and SocialPath – 150. For the PathMMU test tiny split (PathMMU-test-tiny), there are 1,139 questions in total, with the following subset breakdown: Atlas – 208, EduContent – 255, PathCLS – 177, PubMed – 281, and SocialPath – 218. For the PathMMU test split (PathMMU-test), a total of 8,454 questions are included, distributed as: Atlas – 799, EduContent – 1,683, PathCLS – 1,632, PubMed – 2,787, and SocialPath – 1,553.
In addition, we curated pathology-related subsets from established medical benchmarks such as MedXpertQA and OmniMedVQA. From MedXpertQA, 90 pathology cases were selected, whereas the BRIGHT Challenge subset, consisting of 890 cases, was chosen from OmniMedVQA. \\
For the YORN dataset, we selected closed-ended questions from the Path-VQA and Quilt-VQA datasets, with 3,362 and 343 cases respectively. In the MedXpertQA dataset, PathGen-LLaVA-13B ranked first with a score of 25.56, outperforming the tied second-best models Patho-R1-3B and LLaVA-Med-v1.5-Mistral-7B—by 1.12 points. In the BRIGHT Challenge dataset, our reasoning-enhanced pathology model achieved the best performance, surpassing a score of 70 (see Table \ref{mcq3}).
\begin{table}[htbp]
\caption{Comparison of different multimodal models on the pathology subsets of the MedXpertQA and OmniMedVQA benchmarks. The top performance is highlighted in bold, with the second-best underlined.}
\resizebox{\textwidth}{!}{%
\begin{tabular}{rlrr}
\toprule
\multicolumn{2}{c}{} & \multicolumn{1}{c}{\textbf{MedXpertQA}} & \multicolumn{1}{c}{\textbf{OmniMedVQA}} \\
\cmidrule(lr){3-4} 
 & \textbf{Model}  & Path (90) & BRIGHT Challenge (890) \\
 \midrule
\multicolumn{4}{l}{\textbf{Small model without thinking ability}} \\
& Paligemma2-3b-pt-224 & 0 & 1.46  \\
& VILA1.5-3B  & 12.22 & 43.6  \\
& Llama3.2-3B-vl  & 12.22 & 46.4  \\
\midrule
\multicolumn{4}{l}{\textbf{Large model without thinking ability}} \\
& LLaVA-Med-v1.5-Mistral-7B  & \underline{24.44} & 27.75  \\
& HuatuoGPT-Vision-7B  & 12.22 & 52.47  \\
& Quilt-LLaVA-v1.5-7B  & 7.78 & 34.27  \\
& PathGen-LLaVA-13B  & \textbf{25.56} & 46.07  \\
\midrule
\multicolumn{4}{l}{\textbf{Small model with thinking ability}} \\
& DeepSeek-VL2-Tiny & 17.78 & 35.84  \\
& Qwen2.5VL-3B & 23.33 & 55.39  \\
\rowcolor{purple!20} & \textbf{Patho-R1-3B} & \underline{24.44} & \underline{69.89}  \\
\midrule
\multicolumn{4}{l}{\textbf{Large model with thinking ability}} \\
& InternVL2-8B & 10 & 40.56  \\
& InternVL2.5-8B & 22.22 & 49.78  \\
& InternVL3-8B & 15.56 & 65.28  \\
& Llama-3.2-11B-Vision-Instruct & 13.33 & 47.08  \\
& Llama-3.2V-11B-cot & 21.11 & 54.83  \\
& LLaVA-Onevision-7B & 16.67 & 31.46  \\
& Qwen2.5VL-7B & 12.22 & 43.6  \\
\rowcolor{purple!20} & \textbf{Patho-R1-7B} & 22 & \textbf{70.79}  \\
\bottomrule
\end{tabular}%
}
\label{mcq3}
\end{table}

\subsubsection{VQA Benchmark}

To rigorously evaluate the quality of the reasoning steps, we followed the evaluation methodology proposed by LlamaV-O1 \cite{llamavo1} and adopted reasoning steps and final answer correctness as the evaluation metrics. An example of scoring reasoning steps and accuracy is shown below.

\begin{tcolorbox}[
  colback=bodybg,        
  coltext=black,         
  colframe=black,        
  title= Scoring example of accuracy,
  coltitle=white,        
  colbacktitle=titlebg,  
  fonttitle=\bfseries, 
  fontupper=\ttfamily,
  boxrule=0.1mm,         
  arc=2mm,               
  enhanced
]
Message for Deepseek-R1:
\\
"messages": [\{"role": "system", "content": \\
"You are a helpful Assistant. Provide helpful response to the user's question."\}, \\
\{"role": "user", "content": \\"Evaluate the following answer based on accuracy :\\Question : Is interface hepatitis specific to autoimmune hepatitis?\\Ground Truth : Interface hepatitis is not 100\% specific to autoimmune hepatitis. However, in terms of description, the term is somewhat synonymous with autoimmune hepatitis, especially among clinicians who may have a limited understanding of the terminologies used in pathology.\\Model Prediction : Interface hepatitis can occur in autoimmune hepatitis (AIH), but it is not exclusive to AIH. The presence of pan-lobular inflammation with plasma cells and lymphoplasmacytic infiltrates supports AIH, but other conditions like viral hepatitis or drug-induced liver injury must be ruled out. Clinical correlation with serological markers (e.g., autoantibodies) and exclusion of secondary causes are critical for definitive diagnosis\\Match the meaning of the ground truth with the model prediction and if it matches give a 1.Otherwise 0.\\Strictly return only the numeric score , without any additional commentary"\}]
\\
\\
Deepseek-R1's reponse: 
\\
1
\end{tcolorbox}

\begin{tcolorbox}[
  colback=bodybg,        
  coltext=black,         
  colframe=black,        
  title= Scoring example of reasoning steps,
  coltitle=white,        
  colbacktitle=titlebg,  
  fonttitle=\bfseries, 
  fontupper=\ttfamily,
  boxrule=0.1mm,         
  arc=2mm,               
  enhanced
]

\textbf{Message for Deepseek-R1:}
\\
"messages": [\{"role": "system", "content":\\
"You are a reasoning evaluator designed to assess the alignment ,\\coherence , and quality of reasoning steps in text responses .\\Your task is to evaluate reasoning steps between the * ground\\truth * and the * LLM response * using the following metrics : \\

1. ** Faithfulness - Step (1 -10) :**\\- Definition : Measures how well the reasoning steps in the LLM\\response align with the source reasoning steps .\\- Scoring Guidelines :\\- 9 -10: All or almost all steps match or closely reflect the\\ground truth reasoning .\\- 7 -8: Most steps are aligned , with minor deviations .\\- 5 -6: Some steps align , but several are missing or\\significantly altered .\\- 3 -4: Few steps align correctly ; most are off or missing .\\- 1 -2: The majority of steps are not aligned with the source .\\

2. ** Faithfulness - Token (1 -10) :**\\- Definition : Extends Faithfulness - Step to a token - level\\granularity , checking if the content within each reasoning\\step is true to the source .\\- Scoring Guidelines :\\- 9 -10: Token - level details mirror the ground truth closely .\\- 7 -8: Minor token - level deviations but largely faithful .\\- 5 -6: Noticeable inaccuracies in token - level details .\\- 3 -4: Several token - level discrepancies .\\- 1 -2: Most token - level details are incorrect or fabricated .\\

3. ** Informativeness - Step ( Info - Step ) (1 -10) :**\\- Definition : Measures how well the reasoning steps extract all\\relevant information from the source .\\- Scoring Guidelines :\\- 9 -10: Almost all critical information steps are present and\\accurate .\\- 7 -8: Most important points are included , with minor\\omissions .\\- 5 -6: Some key information is missing or underdeveloped .\\- 3 -4: Limited inclusion of critical content .\\- 1 -2: Very poor extraction of relevant information .\\

4. ** Repetition - Token (1 -10) :**\\- Definition : Identifies repeated or unnecessarily paraphrased\\reasoning steps within the hypothesis .\\- Scoring Guidelines :\\- 9 -10: No or minimal unnecessary repetition .\\- 7 -8: Minor repetition that doesn ' t impede clarity .\\- 5 -6: Noticeable repetition that doesn ' t add value .\\- 3 -4: Frequent repetition that disrupts coherence .\\- 1 -2: Excessive repetition reducing the quality of reasoning .\\

5. ** Hallucination (1 -10) :**\\- Definition : Detect irrelevant or invented reasoning steps not\\aligned with the source .\\- Scoring Guidelines :\\- 9 -10: No hallucinations ; all reasoning is grounded in the\\source .

\end{tcolorbox}

\begin{tcolorbox}[
  colback=bodybg,        
  coltext=black,         
  colframe=black,        
  coltitle=white,        
  colbacktitle=titlebg,  
  fonttitle=\bfseries, 
  fontupper=\ttfamily,
  boxrule=0.1mm,         
  arc=2mm,               
  enhanced
]

- 7 -8: One or two minor hallucinations .\\
- 5 -6: Several steps contain invented or irrelevant details .\\
- 3 -4: Many hallucinations, but some grounding remains .\\
- 1 -2: Mostly hallucinated reasoning .\\

6. ** Redundancy (1 -10) :**\\- Definition : Identify redundant reasoning steps that do not add\\value .\\- Scoring Guidelines :\\- 9 -10: No unnecessary steps ; very concise .\\- 7 -8: Minor redundancy .\\- 5 -6: Some steps clearly unnecessary .\\- 3 -4: Many redundant steps .\\- 1 -2: Excessive redundancy that hampers clarity .\\7. ** Semantic Coverage - Step (1 -10) :**\\- Definition : How well the hypothesis covers the essential\\semantic elements from the source reasoning steps .\\- Scoring Guidelines :\\- 9 -10: Almost complete semantic coverage of all important\\elements .\\- 7 -8: Good coverage but some minor elements are missing .\\- 5 -6: Partial coverage with noticeable gaps .\\- 3 -4: Significant semantic gaps .\\- 1 -2: Very poor coverage of essential meaning .\\

8. ** Reasoning Alignment (1 -10) :**\\- Definition : Overall alignment between the hypothesis and the\\reference reasoning chain .\\- Scoring Guidelines :\\- 9 -10: Very closely aligned , minimal divergence .\\- 7 -8: Mostly aligned , with some minor issues .\\- 5 -6: Some alignment , but also several misalignments .\\- 3 -4: Poor alignment , though occasional matches .\\- 1 -2: Fundamentally misaligned reasoning .\\

9. ** Commonsense (1 -10) :**\\- Definition : Check for missing commonsense reasoning required\\to solve the problem .\\- Scoring Guidelines :\\- 9 -10: Adequate commonsense reasoning present .\\- 7 -8: Minor commonsense gaps but mostly adequate .\\- 5 -6: Noticeable commonsense gaps .\\- 3 -4: Many commonsense steps missing .\\- 1 -2: Almost entirely lacking necessary commonsense .\\

10. ** Missing Step (1 -10) :**\\- Definition : Identify if any necessary reasoning steps are\\missing .\\- Scoring Guidelines :\\- 9 -10: No critical steps missing .\\- 7 -8: Minor missing steps that don ' t significantly affect\\the conclusion .\\- 5 -6: Some important steps absent , affecting the outcome .\\- 3 -4: Several crucial missing steps .\\- 1 -2: Major gaps ; the reasoning chain is incomplete .\\

** Additional Instructions for Consistency :**\\- Always follow the above scoring guidelines strictly .\\

\end{tcolorbox}

\begin{tcolorbox}[
  colback=bodybg,        
  coltext=black,         
  colframe=black,        
  coltitle=white,        
  colbacktitle=titlebg,  
  fonttitle=\bfseries, 
  fontupper=\ttfamily,
  boxrule=0.1mm,         
  arc=2mm,               
  enhanced
]

- Before scoring , re - read both the ground truth and the LLM\\response carefully .
\\- Compare the reasoning steps directly to determine where they\\align or diverge .\\- Use the provided scoring benchmarks ( anchor examples , if any ) as\\a reference to maintain consistency across evaluations .\\- Avoid subjective interpretation and adhere to the given\\thresholds .\\- Once scores for all metrics are determined , compute the Overall\\Score as the average of all metric scores .\\
- Provide the final output as a Python dictionary with the\\structure only don ' t add a anything extra , beacuase your out\\will be used in code pipeline . So single change in you output\\will crash whole system . :\\

Example output : \{ 'Faithfulness - Step': 8.0 , 'Faithfulness - Token':7.5 , 'Informativeness - Step': 8.5 , 'Repetition - Token': 9.0 ,'Hallucination': 9.5 , ' Redundancy': 8.0 , 'Semantic Coverage - Step': 8.5 , 'Reasoning Alignment': 8.0 , 'Commonsense': 9.0 , 'Missing Step': 8.5 , 'Overall Score': 8.65\}\\

Do not give output in following format :'''python \{' Faithfulness - Step ': 1.0 ,' Faithfulness - Token ': 1.0 ,' Informativeness - Step ': 1.0 ,' Repetition - Token ': 9.0 , ' Hallucination ': 1.0 ,' Redundancy ': 9.0 ,' Semantic Coverage - Step ': 1.0 ,' Reasoning Alignment ': 1.0 ,' Commonsense ': 1.0 ,' Missing Step ': 1.0 ,' Overall Score ': 2.6\}'''\} \\

\{"role": "user", "content": "\\        
Question: Is interface hepatitis specific to autoimmune hepatitis? \\ 
Ground Truth : Interface hepatitis is not 100\% specific to autoimmune hepatitis. However, in terms of description, the term is somewhat synonymous with autoimmune hepatitis, especially among clinicians who may have a limited understanding of the terminologies used in pathology.\\
LLM Response: The image shows portal inflammation with interface hepatitis, a feature commonly seen in autoimmune hepatitis (AIH). However, interface hepatitis is not pathognomonic for AIH and can occur in other conditions like viral hepatitis or drug-induced liver injury. In AIH, the inflammation typically involves plasma cells and lymphocytes at the portal-parenchymal interface, accompanied by lobular activity and hepatocyte injury. While the presence of interface hepatitis supports AIH, it must be correlated with other histological features (e.g., plasma cell-rich infiltrate, emperipolesis) and serological markers (e.g., anti-nuclear antibodies, anti-smooth muscle antibodies) to exclude mimics. The absence of viral inclusions or granulomas further narrows the differential"\}]
\\

\textbf{Deepseek-R1's response:}\\
\{'Faithfulness - Step': 8.0, 'Faithfulness - Token': 7.5, 'Informativeness - Step': 8.5, 'Repetition - Token': 9.0, 'Hallucination': 9.5, 'Redundancy': 8.0, 'Semantic Coverage - Step': 8.5, 'Reasoning Alignment': 8.0, 'Commonsense': 9.0, 'Missing Step': 8.5, 'Overall Score': 8.45\}
\end{tcolorbox}

Since our evaluation includes the reasoning process, we selected models that adhere to the <think></think> <answer></answer> output format as required by the prompt. Models that fail to follow this format, such as LLaVA-Med and HuatuoGPT-Vision, were excluded from our evaluation.

\begin{tcolorbox}[
  colback=bodybg,        
  coltext=black,         
  colframe=black,        
  title= Example outputs from models that fail to follow the instructed format,
  coltitle=white,        
  colbacktitle=titlebg,  
  fonttitle=\bfseries, 
  fontupper=\ttfamily,
  boxrule=0.1mm,         
  arc=2mm,               
  enhanced
]

Question: what does the trabecular bone forming the marrow space show at the margins?
 \\
 
Ground truth: trabeculae with osteoclastic activity
\\

Prompt: You are a pathology expert, your task is to answer question step by step. Use the following format:<think> Your step-by-step reasoning </think><answer> Your final answer </answer>
\\

LLaVA-Med's Output: The trabecular bone forming the marrow space shows a thin rim of bone at the margins. This observation is made using a light microscope at a magnification of 100x.
\\

HuatuoGPT-vision's Output: 1. The image shows a cross-section of trabecular bone forming the marrow space. 2. The trabecular bone appears to have a honeycomb-like structure with interconnected spaces. 3. The margins of the trabecular bone show a distinct pattern of bone formation. 4. The bone formation at the margins is likely due to the activity of osteoblasts, which are responsible for new bone formation. 5. The presence of osteoblasts at the margins indicates active bone remodeling and growth. 6. The trabecular bone structure and the activity at the margins suggest a healthy bone remodeling process. <answer> The trabecular bone forming the marrow space shows active bone remodeling and growth at the margins, with the presence of osteoblasts.

\end{tcolorbox}

The accuracy results are presented in the experiments part of the paper. We conducted a comprehensive evaluation of multimodal reasoning quality from two complementary perspectives: Chain-of-Thought and Chain-of-Draft. The evaluation covers ten fine-grained metrics, including Commonsense, Faithfulness, Hallucination, Informativeness, Reason Alignment, Redundancy, Repetition, and Semantic Coverage. The Patho-R1 model demonstrates strong and consistent performance across most of these dimensions, particularly excelling in Commonsense, Faithfulness, and Reason Alignment. InternVL2.5-8B achieves the best result in the Repetition metric, indicating a lower tendency to produce redundant output, while LLaMA-3.2-11B-Vision-Instruct achieves the highest score in Semantic Coverage, reflecting its comprehensive understanding of input content.

Detailed metric-wise evaluations of the reasoning steps, for models conforming to the required output format, are reported in Table~\ref{quilt-open} and Table~\ref{path-open}, which correspond to the open-ended Quilt-VQA and Path-VQA datasets, respectively.

\begin{table}[h]
\caption{Comparison of multimodal models on reasoning-related metrics on Quilt-VQA benchmarks. Each cell reports CoT/CoD results. The top performance is highlighted in bold, with the second-best underlined. Metrics: Faith. = Faithfulness, Halluc. = Hallucination, Info. = Informativeness, Redund. = Redundancy, Rep. = Repetition,
Sem. Cov. = Semantic Coverage.}
\renewcommand{\arraystretch}{1.3}
\resizebox{\textwidth}{!}{%
\begin{tabular}{lccccccccccc}
\toprule
 \textbf{Model} & \textbf{Commonsense} & \textbf{Faith.-Step} & \textbf{Faith.-Token} & \textbf{Halluc.} & \textbf{Info.-Step} & \textbf{Miss Step} & \textbf{Reason Align.} & \textbf{Redund.} & \textbf{Rep.-Token} & \textbf{Sem. Cov.-Step} & \textbf{Overall} \\
\midrule
InternVL2-8B & 4.3/3.35 & 2.51/1.95 & 2.42/1.88 & 2.78/2.44 & 3.03/2.11 & 2.97/2.15 & 3.94/3.49 & 2.52/1.95 & 7.48/\underline{8.6} & 8.58/\underline{9.02} & 2.9/2.05 \\
InternVL2.5-8B & 2.72/2.86 & 2.61/1.76 & \underline{2.58}/1.71 & 2.7/2.15 & 2.57/1.94 & 2.59/1.95 & 3.92/3.32 & 2.59/1.76 & \textbf{9.16}/8.58 & \textbf{9.16}/9 & 2.58/1.88 \\
InternVL3-8B & 3.83/4.09 & 2.61/2.37 & 2.52/2.28 & \underline{2.79}/2.75 & 2.96/2.68 & 2.95/2.73 & \underline{3.98}/3.86 & 2.61/2.37 & 8/8.31 & 8.73/8.9 & 2.85/2.58 \\
Llava-Onevision-7B & 2.1/1.82 & 1.63/1.43 & 1.59/1.39 & 1.69/1.48 & 1.73/1.48 & 1.72/1.49 & 3.1/2.94 & 1.62/1.41 & 8.46/8.59 & 8.88/8.95 & 1.69/1.46 \\
LLaMA-3.2-11B-Vision-Instruct & 2.76/2.69 & 2.53/2.45 & 2.5/2.42 & 2.67/2.57 & 2.52/2.43 & 2.54/2.44 & 3.87/3.79 & 2.52/2.43 & \underline{9.1}/\textbf{9.09} & \underline{9.14}/\textbf{9.13} & 2.52/2.43 \\
LLaMA-3.2V-11B-CoT & 3.54/3.46 & 1.98/1.99 & 1.89/1.88 & 2.62/2.66 & 2.3/2.26 & 2.39/2.38 & 3.56/3.56 & 1.99/1.99 & 7.96/8.05 & 8.97/8.96 & 2.22/2.19 \\
   \rowcolor{purple!20}
Patho-R1-3B & \textbf{5.14}/\underline{4.69} & \textbf{3.1}/\underline{2.84} &\textbf{ 2.75}/\underline{2.6} & \textbf{3.04}/\underline{2.78} & \textbf{3.77}/\underline{3.45} & \textbf{3.57}/\underline{3.33} & \textbf{4.27}/\underline{4.07} & \textbf{3.07}/\underline{2.83} & 6.74/6.69 & 8.17/8.37 & \textbf{3.42}/\underline{3.17} \\
   \rowcolor{purple!20}
Patho-R1-7B & \underline{4.6}/\textbf{5.05} & \underline{2.72}/\textbf{3.01} & 2.5/\textbf{2.78} & 2.7/\textbf{3.01} & \underline{3.33}/\textbf{3.71} & \underline{3.23}/\textbf{3.59} & 3.97/\textbf{4.22} & \underline{2.72}/\textbf{3.02} & 6.65/6.39 & 8.32/8.27 & \underline{3.06}/\textbf{3.41} \\
\bottomrule
\end{tabular}%
}
\label{quilt-open}
\end{table}
\begin{table}[h]
\caption{Comparison of multimodal models on reasoning-related metrics on Path-VQA benchmarks. Each cell reports CoT/CoD results. The top performance is highlighted in bold, with the second-best underlined. Metrics: Faith. = Faithfulness, Halluc. = Hallucination, Info. = Informativeness, Redund. = Redundancy, Rep. = Repetition,
Sem. Cov. = Semantic Coverage.}
\renewcommand{\arraystretch}{1.3}
\resizebox{\textwidth}{!}{%
\begin{tabular}{lccccccccccc}
\toprule
 \textbf{Model} & \textbf{Commonsense} & \textbf{Faith.-Step} & \textbf{Faith.-Token} & \textbf{Halluc.} & \textbf{Info.-Step} & \textbf{Miss Step} & \textbf{Reason Align.} & \textbf{Redund.} & \textbf{Rep.-Token} & \textbf{Sem. Cov.-Step} & \textbf{Overall} \\
\midrule
 InternVL2-8B & 1.65/1.58 & 1.33/1.31 & 1.32/1.30 & 1.37/1.39 & 1.43/1.36 & 1.41/1.37 & 1.33/1.31 & 8.63/8.86 & 8.93/\underline{9.01} & 1.40/1.34 & 2.88/2.87 \\
 InternVL2.5-8B & 1.36/2.05 & 1.36/1.46 & 1.36/1.43 & 1.36/1.59 & 1.36/1.60 & 1.36/1.58 & 1.36/1.46 & \textbf{9.04}/8.44 & \textbf{9.04}/9.00 & 1.36/1.55 & 2.90/3.01 \\
 InternVL3-8B & 1.57/2.25 & 1.44/1.70 & 1.43/1.66 & 1.46/\underline{1.80} & 1.48/1.84 & 1.48/1.86 & 1.44/1.69 & 8.90/8.62 & 8.98/8.93 & 1.47/1.79 & 2.97/\underline{3.21} \\
 Llava-Onevision-7B & 1.62/1.28 & 1.43/1.15 & 1.42/1.15 & 1.45/1.18 & 1.49/1.16 & 1.48/1.17 & 1.43/1.15 & 8.80/\underline{8.96} & 8.97/9.00 & 1.47/1.16 & 2.95/2.73 \\
 Llama-3.2-11B-Vision-Instruct & 1.43/\textbf{2.69} & 1.42/\textbf{2.45} & 1.42/\textbf{2.42} & 1.42/\textbf{2.57} & 1.42/\textbf{2.43} & 1.42/\textbf{2.44} & 1.41/\textbf{2.43} & \textbf{9.04}/\textbf{9.09} & \textbf{9.04}/\textbf{9.13} & 1.42/\textbf{2.43} & 2.94/\textbf{3.79} \\
 Llama-3.2V-11B-cot & 1.66/1.93 & 1.43/1.49 & 1.42/1.46 & 1.48/1.62 & 1.49/1.59 & 1.50/1.61 & 1.43/1.49 & 8.76/8.72 & 8.98/8.99 & 1.48/1.55 & 2.96/3.04 \\
   \rowcolor{purple!20}
     Patho-R1-3B & \underline{2.19}/2.15 & \underline{1.66}/1.65 & \underline{1.62}/1.61 & \underline{1.66}/1.65 & \underline{1.84}/1.83 & \underline{1.79}/1.78 & \underline{1.66}/1.65 & 8.31/8.34 & 8.81/8.84 & \underline{1.77}/1.76 & \underline{3.13}/3.12 \\
   \rowcolor{purple!20}
     Patho-R1-7B & \textbf{2.37}/\underline{2.38} & \textbf{1.82}/\underline{1.77} &\textbf{ 1.77}/\underline{1.71} & \textbf{1.83}/1.77 & \textbf{2.02}/\underline{1.99} & \textbf{1.96}/\underline{1.93} & \textbf{1.82}/\underline{1.77} & 8.20/8.05 & 8.79/8.77 & \textbf{1.94}/\underline{1.89} & \textbf{3.25}/3.20 \\
\bottomrule
\end{tabular}%
}
\label{path-open}
\end{table}

\subsection{Ablation Experiment} \label{appendix:ablation experiment}

We conducted two primary ablation studies to investigate: (1) the impact of model size and training stage (Base / CPT\&SFT / GRPO or DAPO for both 3B and 7B models), and (2) the effect of training steps on final performance.

First, we evaluated the 3B and 7B versions of Qwen2.5VL on the PathMMU-test-tiny benchmark at different stages of training. Specifically, we tested the base model, then re-evaluated it after the Completion Prompt Tuning and Supervised Fine-tuning stages. Finally, we assessed performance after applying GRPO and DAPO reinforcement learning techniques.

Experimental results show that 3B model, after completing all three stages of training, achieved an overall improvement of approximately 20\% over the base model. The 7B model exhibited an even more significant performance gain of around 27\% (see Table \ref{ablation37bstage}). In our final model selection, we adopt DAPO for the 3B variant (Patho-R1-3B) and GRPO for the 7B variant (Patho-R1-7B). This decision is primarily driven by overall experimental results, where each optimization strategy demonstrated the most consistent and robust performance for its corresponding model size across a wide range of benchmarks.

\begin{table}[htbp]
\caption{Ablation results of Qwen2.5-VL-3B and Qwen2.5-VL-7B on the Pathmmu-test-tiny benchmark, showing the effects of CPT\&SFT and GRPO/DAPO across five source-specific subsets.}
\centering
\setlength{\tabcolsep}{4pt}
\renewcommand{\arraystretch}{1.3}
\resizebox{\textwidth}{!}{
\begin{tabular}{lcccccc}
\toprule
\textbf{Variants} & Atlas (208) & EduContent (255) & PathCLS (177) & PubMed (281) & SocialPath (218) & Overall \\
\midrule
Base (3B) & 46.63 & 44.31 & 23.73 & 49.82 & 46.79 & 43.37 \\
\rowcolor{blue!10}
+CPT\&SFT (Stage1/2) & 70.19 & 67.06 & 25.42 & 61.92 & 61.47 & 58.82 \\
\rowcolor{blue!10}
+Stage1/2\&(GRPO/DAPO) & 75.48 / 74.52 & 68.63 / 67.45 & 25.99 / 36.72 & 66.90 / 67.62 & 64.68 / 66.51 & 62.07 / 63.83 \\
\rowcolor{green!10}
$\Delta$ & +28.85 / +27.89 & +24.32 / +23.14 & +2.26 / +12.99 & +17.08 / +17.80 & +17.89 / +19.72 & +18.70 / +20.46 \\
\midrule
Base (7B) & 44.23 & 49.41 & 24.86 & 44.84 & 40.83 & 41.88 \\
\rowcolor{blue!10}
+CPT\&SFT (Stage1/2) & 73.08 & 69.02 & 38.42 & 66.19 & 62.84 & 63.13 \\
\rowcolor{blue!10}
+Stage1/2\&(GRPO/DAPO) & 81.73 / 82.69 & 75.29 / 72.55 & 44.63 / 41.24 & 72.24 / 72.24 & 67.89 / 69.72 & 69.53 / 68.92 \\
\rowcolor{green!10}
$\Delta$ & +37.50 / +38.46 & +25.88 / +23.14 & +19.77 / +16.38 & +27.40 / +27.40 & +27.06 / +28.89 & +27.65 / +27.04 \\
\bottomrule
\end{tabular}
}
\label{ablation37bstage}
\end{table}

\begin{wrapfigure}{l}{0.75\linewidth}
    \centering
    \includegraphics[width=\linewidth]{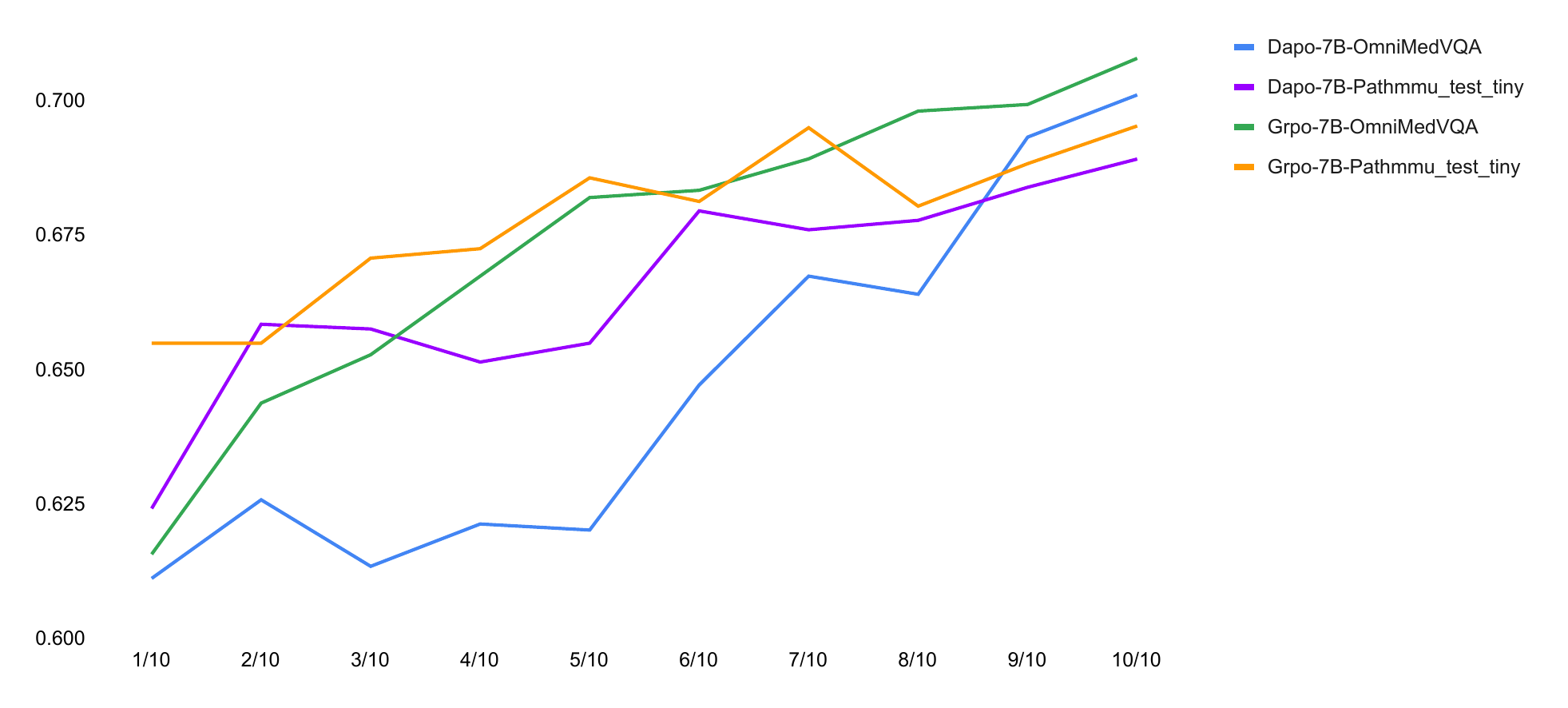}
    \caption{Ablation study comparing GRPO and DAPO on OmniMedVQA and Pathmmu-test-tiny. The x-axis shows the proportion of total training steps completed, and the y-axis indicates model accuracy.}
    \label{ablation7b}
\end{wrapfigure}

We further investigated how the best-performing 7B model behaves under different reinforcement learning strategies (GRPO vs. DAPO) and varying numbers of training steps. Results demonstrate that on the larger PathMMU-test-tiny dataset, both GRPO and DAPO led to notable improvements and faster convergence. Among them, GRPO showed a slightly better overall performance, suggesting it is more effective in optimizing model reasoning and decision-making capabilities in the pathology domain (see Figure \ref{ablation7b}).

\section{Broader Impacts}

The development of Patho-R1 marks a significant step forward in advancing multimodal artificial intelligence in the field of pathology, a domain that has long been underrepresented in vision-language research. By leveraging domain-specific resources, including expert-curated textbooks and authentic clinical reasoning from practicing pathologists, we bridge a critical gap between general-purpose medical vision-language models and the complex demands of pathology. This work carries profound implications for clinical practice, scientific research, and the broader landscape of AI-driven medical innovation.

Patho-R1 focuses on perception-driven learning and introduces a rule-based reinforcement learning framework that enhances the accuracy and interpretability of AI systems in pathological image analysis. This is particularly crucial in high-stakes medical settings where accurate diagnoses hinge on subtle visual cues. By prioritizing fine-grained perception over purely high-level reasoning, Patho-R1 paves the way for more reliable, transparent, and clinically actionable AI tools. These tools have the potential to assist pathologists in diagnosing complex diseases, reduce diagnostic errors, and improve patient outcomes—especially in underserved areas where access to expert pathologists is limited.

Overall, this research contributes to the long-term goal of building transparent, explainable, and clinically valuable AI systems in pathology, while also laying the groundwork for multimodal learning in other complex areas of medicine.

\end{document}